%% file: arxiv.tex
\documentclass{article} 
\usepackage{iclr2026_conference,times}

\input{math_commands}

\usepackage{hyperref}
\usepackage{url}
\usepackage{graphicx}
\usepackage{soul}
\usepackage{booktabs}
\usepackage{multicol}
\usepackage{multirow}
\usepackage[table]{xcolor}
\usepackage{caption}
\usepackage{wrapfig}
\usepackage{makecell}
\usepackage{graphicx}
\usepackage{enumitem}
\usepackage{listings}
\iclrfinalcopy

\definecolor{blue}{HTML}{8eb4db}
\definecolor{pink}{HTML}{e0bcb2}
\definecolor{FIRST}{HTML}{d4e9d8}
\definecolor{SECOND}{HTML}{fdf3d0}

\def \eg {{\emph{e.g.}\thinspace}}

\title{EgoHandICL: Egocentric 3D Hand Reconstruction with In-Context Learning}

\author{Binzhu Xie$^{1,2}$\footnotemark[1]~~, Shi Qiu$^{1,2}$\footnotemark[1]~~\footnotemark[2]~~, Sicheng Zhang$^{3}$, Yinqiao Wang$^{1,2}$, Hao Xu$^{1,2}$,\\ 
\textbf{Muzammal Naseer$^{3}$, Chi-Wing Fu$^{1,2}$, Pheng-Ann Heng$^{1,2}$} \\
$^1$ Department of Computer Science and Engineering, The Chinese University of Hong Kong \\
$^2$ Institute of Medical Intelligence and XR, The Chinese University of Hong Kong \\
$^3$ Department of Computer Science, Khalifa University\\
\texttt{\{bzxie,shiqiu\}@cse.cuhk.edu.hk}
}

\renewcommand{\thefootnote}{\fnsymbol{footnote}}
\footnotetext[1]{Equal contribution.}
\footnotetext[2]{Corresponding author.}
\renewcommand*{\thefootnote}

\begin{document}

\maketitle
\begin{figure}[htbp]
\centering
\includegraphics[width=\linewidth]{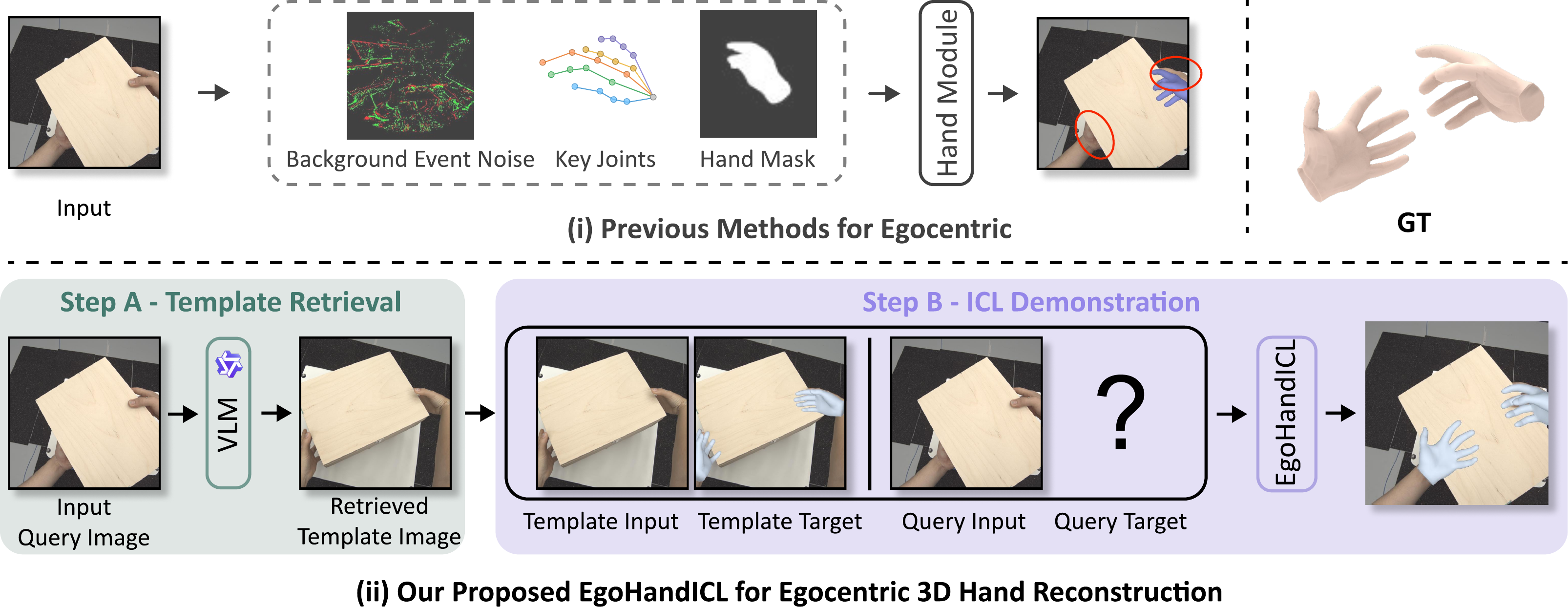}
\caption{
\textbf{Comparing EgoHandICL with previous methods.} (i) Prior works improve egocentric hand reconstruction by exploiting auxiliary cues~\citep{prakash20243d,Hara2025EventEgoHands}, thereby having limited capabilities in handling challenging scenarios with severe occlusions. (ii) EgoHandICL improves the precision through two key steps. \textbf{Step A:} We prompt vision-language models with egocentric cues to retrieve template images for the input query image. \textbf{Step B:} We construct ICL demonstrations by aligning the input-target pairs of both template and query images.}
\label{fig:teaser}
\end{figure}

\begin{abstract}
Robust 3D hand reconstruction is challenging in egocentric vision due to depth ambiguity, self-occlusion, and complex hand-object interactions.
Prior works attempt to mitigate the challenges by scaling up training data or incorporating auxiliary cues, often falling short of effectively handling unseen contexts.
In this paper, we introduce \textbf{EgoHandICL}, the first in-context learning (ICL) framework for 3D hand reconstruction that achieves strong semantic alignment, visual consistency, and robustness under challenging egocentric conditions. 
Specifically, we develop (i) complementary exemplar retrieval strategies guided by vision–language models (VLMs), (ii) an ICL-tailored tokenizer that integrates multimodal context, and (iii) a Masked Autoencoders (MAE)-based architecture trained with 3D hand–guided geometric and perceptual objectives.
By conducting comprehensive experiments on the ARCTIC and EgoExo4D benchmarks, our EgoHandICL consistently demonstrates significant improvements over state-of-the-art 3D hand reconstruction methods.
We further show EgoHandICL’s applicability by testing it on 
real-world egocentric cases and integrating it with EgoVLMs to enhance their hand–object interaction reasoning.
Our code and data are available at: \url{https://github.com/Nicous20/EgoHandICL}.
\end{abstract}
\input{sec/1_intro}
\input{sec/2_related_work}
\input{sec/3_method}
\input{sec/4_exp}

\input{sec/5_application}

\input{sec/6_conclusion}




\bibliography{iclr2026_conference}
\bibliographystyle{iclr2026_conference}

\newpage
\appendix
\input{sec/appendix}

\end{document}

%% file: math_commands.tex
\usepackage{amsmath,amsfonts,bm}

\def\eqref#1{equation~\ref{#1}}

\def\1{\bm{1}}

\DeclareMathAlphabet{\mathsfit}{\encodingdefault}{\sfdefault}{m}{sl}
\SetMathAlphabet{\mathsfit}{bold}{\encodingdefault}{\sfdefault}{bx}{n}



%% file: sec/1_intro.tex
\section{Introduction}
Reconstructing 3D hands from monocular RGB images has long been a core computer vision task, with a wide range of applications in extended reality (XR), human–computer interaction (HCI), robotics, \emph{etc}. 
By leveraging transformer-based backbones trained on large-scale datasets to extract rich visual representations~\citep{cui2023adaptive}, recent methods, such as HaMeR~\citep{pavlakos2024reconstructing} and WiLoR~\citep{potamias2025wilor}, achieve strong 3D hand reconstruction performance across various benchmarks. 
While effective, these models can still encounter practical challenges, such as depth ambiguity, self-occlusion, and domain shift, which often affect their robustness and generalization in real-world applications~\citep{hong2022pointcam,zhu2024ssp,cui2024numgrad}.
These challenges are further aggravated in egocentric settings, where severe occlusions, perspective distortions, and complex hand–object interactions are commonly present.

Previous works~\citep{prakash20243d,Hara2025EventEgoHands} have also explored specialized solutions tailored for egocentric 3D hand reconstruction: as illustrated in Fig.~\ref{fig:teaser}-(i), current methods 
utilize auxiliary supervision cues that require additional annotations, but still fail to resolve severe occlusions or ambiguous hand-object interactions in egocentric views. 
These limitations highlight the urgent need for a flexible and generalizable framework that can adapt to diverse egocentric contexts with robustness.
Although each egocentric scenario presents unique visual ambiguities, humans naturally resolve them by drawing on prior experience, multimodal context, and task-specific knowledge. This remarkable ability inherently aligns with the core concept of in-context learning (ICL)~\citep{NEURIPS2020_1457c0d6}, which adapts to solve new problems by conditioning on a few relevant contextual examples. While ICL has achieved remarkable progress in language modeling~\citep{dong2022survey,ferber2024context}, recent studies have begun to apply this paradigm for vision tasks, where rapid problem adaptation and example-guided reasoning are equally critica~\citep{zong2024vlicl}. 
Given that ICL mirrors how humans exploit relevant cues to overcome ambiguity, it provides a natural paradigm for egocentric vision. Motivated by this inherent connection, we formulate a new framework, namely EgoHandICL, to advance egocentric 3D hand reconstruction by exploiting the strong reasoning capabilities of the in-context learning paradigm.
As the first ICL-based approach for 3D hand reconstruction, we address the following two critical issues.

First, ICL’s effectiveness highly depends on selecting relevant examples~\citep{liu-etal-2022-makes,NEURIPS2023_398ae57e,rubin-etal-2022-learning}, yet egocentric scenarios make this selection particularly difficult. 
As shown in Fig. \ref{fig:teaser}-(ii), we propose complementary retrieval strategies for exemplar template selection: (i) using four hand-involvement modes (left-hand, right-hand, two-hand, and non-hand) that broadly cover typical egocentric hand-related activities, we retrieve examples with similar hand involvement to ensure visual consistency between the query image and the matched templates; and (ii) we design interaction-specific prompts with a vision–language model (VLM) to retrieve adaptive templates conditioned on semantic context. 
Together, these strategies ensure that exemplars with strong semantic alignment and visual fidelity are exploited during in-context learning.

Second, unlike conventional ICL methods addressing single-modal mappings, 3D hand reconstruction requires bridging 2D visual inputs and 3D parametric outputs. 
To ensure structural consistency between queries and templates, we represent both inputs and outputs using a unified MANO parameterization~\citep{romero2017embodied}, which yield semantically aligned input–output pairs that facilitate effective ICL. 
Moreover, we incorporate egocentric hand priors and multimodal cues to encode visual, textual, and structural context into a learnable token space, and adopt a Mask Autoencoders (MAE)-based architecture~\cite{he2022masked} trained with 3D geometric and perceptual objectives. 
By conducting extensive experiments, we demonstrate that these designs enable our EgoHandICL framework to achieve more robust and generalizable egocentric hand reconstruction performance. In summary, our main contributions are threefold:
\setlist{nolistsep}
\begin{itemize}
[noitemsep,leftmargin=*] 
\item We present the first in-context learning approach to 3D hand reconstruction, showcasing a promising way to tackle severe occlusions and diverse interactions in egocentric vision. 
\item We develop the EgoHandICL framework, which retrieves effective exemplars, tokenizes multimodal context, and 
learns through an MAE-style architecture
for robust and generalizable 3D hand reconstruction. 
\item We present extensive experiments, demonstrating that our EgoHandICL achieves state-of-the-art performance on egocentric benchmarks and exhibits strong practicality in real-world scenarios, including self-captured data and enhancing hand–object interaction reasoning in VLMs.
\end{itemize}

%% file: sec/2_related_work.tex
\section{Related Work}

\noindent \textbf{3D Hand Reconstruction.}
Hand reconstruction has been extensively studied for a long time, where early works used to employ depth cameras to recover 3D hand joints and meshes~\citep{ge2016robust,oikonomidis2011efficient,tagliasacchi2015robust,rogez20143d,simon2017hand,sridhar2016real,sun2015cascaded,tompson2014real}. A milestone is the introduction of MANO~\citep{romero2017embodied}, a low-dimensional parametric hand model, which enables single-image 3D hand reconstruction by regressing associated posed shapes. \cite{boukhayma20193d} then demonstrates the applicability of MANO by using a CNN model and an articulated hand mesh. This learning-based approach has inspired following methods that either directly regress MANO parameters~\citep{baek2019pushing,potamias2023handy,baek2020weakly} or predict hand vertices for improved image alignment~\citep{kulon2019single,choi2020pose2mesh,ge20193d,kulon2020weakly}. Beyond parametric regression, alternative strategies have been explored, including voxel-based~\citep{iqbal2018hand,moon2020i2l} and vertex regression methods~\citep{chen2021camera}, with a few work further emphasizing efficiency and robustness of reconstruction~\citep{chen2022mobrecon,park2022handoccnet,oh2023recovering,jiang2023probabilistic}. More recently, large-scale vision transformers pretrained on millions of images demonstrate that scaling both model capacity and data volume significantly enhances the generalization of hand reconstruction models to in-the-wild scenarios~\citep{dong2024hamba,kim2023sampling,lin2021end}. In particular, HaMeR~\citep{pavlakos2024reconstructing} presents a powerful vision transformer-based~\citep{vaswani2017attention} pipeline that utilizes patched image tokens to reconstruct the MANO parametric model, achieving state-of-the-art accuracy in both egocentric and in-the-wild settings. However, existing methods mainly focus on general reconstruction and lack robustness under challenging interactions and occlusions: \eg, when hands cross and one is heavily obscured (Fig.~\ref{fig:arctic}), or when a hand blends into the background due to a black glove (Fig.~\ref{fig:egoexo}). State-of-the-art approaches tend to miss hands, confuse left and right identities, or distort occluded regions. In contrast, our method leverages multimodal cues and acquires in-context knowledge from retrieved exemplars, achieving consistent and accurate hand reconstruction in these difficult egocentric scenarios.

\noindent \textbf{In-Context Learning in Vision.}
Popularized by GPT-3~\citep{NEURIPS2020_1457c0d6}, in-context learning (ICL) allows models to adapt to new tasks by conditioning on a few input–output demonstrations without updating model parameters~\citep{pmlr-v139-radford21a,rubin-etal-2022-learning,xie2021explanation}. Inspired by the success of ICL in LLM exploration, researchers have also extended this paradigm to computer vision research, with representative efforts on segmentation~\citep{li2024visual}, recognition~\citep{zhang2025contextdrivingincontextlearning}, and multimodal understanding~\citep{zong2024vlicl}.
However, applying ICL to 3D vision is more challenging because 3D data involves complex spatial and temporal structures that are not easily captured or represented by simple input–output pairs. Consequently, only a few
studies have explored this area, such as PIC~\citep{fang2023explore} for point cloud recognition, HiC~\citep{liu2025human} for human motion, and TrajICL~\citep{fujii2025towards} for pedestrian trajectory forecasting, \emph{etc}.
Yet, none of them learns the contextual information needed to handle the large modality gap between 2D images and 3D meshes that exists in the reconstruction problem.
To this end, we unify both modalities in the MANO parameter space and develop an ICL-specific tokenizer to effectively incorporate multimodal context. Moreover, within our EgoHandICL framework, the proposed VLM-guided exemplar retrieval strategies and MAE-driven reconstruction pipeline together facilitate semantic alignment and geometric consistency, enabling robust egocentric hand reconstruction even under severe occlusion and visual ambiguities.

%% file: sec/3_method.tex
\section{Method}
In this section, we first formulate in-context learning for egocentric 3D hand reconstruction  (Sec.~\ref{sec:icl_modeling}). 
Then, we present our EgoHandICL framework (Sec.~\ref{sec:framework}), including template retrieval strategies that select exemplar samples, an ICL tokenizer that integrates visual, textual, and structural context, as well as the training and inference in practice. 
Finally, we introduce the specific losses that are used to train the EgoHandICL framework (Sec.~\ref{sec:loss}). 

\subsection{Modeling In-Context Learning in Egocentric 3D Hand Reconstruction}
\label{sec:icl_modeling}

\noindent \textbf{ICL Preliminaries.} 
In-context learning (ICL) typically models a task as few-shot inference, where a model $\mathcal{F}$ conditions on a small set of input–target exemplars and a query within a shared context. 
Formally, given a query \( x^{\text{qry}} \) and a context set \( \mathcal{C} = \{(x_i, y_i)\}_{i=1}^N \) of $N$ task-specific exemplar pairs (\emph{i.e.}, ``templates''), the model predicts \( y^{\text{qry}} \) by conditioning on both the query and the templates:
\begin{equation}
    y^{\text{qry}} = \mathcal{F}(x^{\text{qry}} | \mathcal{C}).
    \label{eq:icl}
\end{equation}

\noindent \textbf{3D Hand Reconstruction.} 
Given an image \(I \in \mathbb{R}^{H \times W \times 3}\), we aim to reconstruct the 3D mesh of each hand \(i\) represented by the MANO model~\citep{romero2017embodied}. 
We denote the set of MANO hand parameters as \(\mathcal{M} = \{\Theta_i, \beta_i, \Phi_i\}_{i=1}^N\), where \(\Theta_i \in \mathbb{R}^{15 \times 3}\) are the pose parameters, \(\beta_i \in \mathbb{R}^{10}\) are the shape parameters, and \(\Phi_i \in \mathbb{R}^3\) is the global orientation. 
Our objective is to learn a mapping:
\begin{equation}
    \mathcal{M} = \mathcal{G}(I),
    \label{eq:hand}
\end{equation}
where \(\mathcal{G}\) predicts the MANO parameters for all visible hands in the given image $I$. 
The estimated parameters \( \mathcal{M} \) are then passed to the MANO decoder to obtain the corresponding 3D hand meshes. 

\noindent \textbf{ICL for Egocentric 3D Hand Reconstruction.} 
Although we can realize 3D hand reconstruction by directly regressing MANO parameters from a single image, it becomes far more difficult in egocentric views, which often suffer from occlusions, complex hand–object interactions, and ambiguous viewpoints. 
To mitigate this, we leverage ICL with dedicated egocentric contextual exemplars, enabling example-based reasoning for accurate and robust egocentric 3D hand reconstruction. 
To fulfill the objective of Eq.~\ref{eq:hand} within the ICL formulation of Eq.~\ref{eq:icl}, we proceed as follows. 

Given a query image $I_\text{qry}$, we first retrieve template images that are contextually related to the query. 
For each retrieved template image \(I_{\text{tpl}}\), we construct an input–target exemplar pair $(\tilde{\mathcal{M}}_{\text{tpl}}, \mathcal{M}_{\text{tpl}})$, aligning with both the objective of the 3D hand reconstruction task as well as the requirement of the ICL paradigm. 
Particularly, \(\tilde{\mathcal{M}}_{\text{tpl}}\) denotes coarse MANO parameters estimated by applying a trained reconstruction model (\emph{e.g.}, HaMeR~\citep{pavlakos2024reconstructing} or WiLoR~\citep{potamias2025wilor}) to \(I_{\text{tpl}}\), as in Eq.~\ref{eq:hand}; while \(\mathcal{M}_{\text{tpl}}\) denotes the retrieved template's ground-truth MANO parameters from the database. 
Then, we obtain a context set of egocentric exemplars $\mathcal{C}_{\mathcal{M}}=\{(\tilde{\mathcal{M}}_{\text{tpl}}, \mathcal{M}_{\text{tpl}})\}$. 
Finally, we estimate the coarse MANO parameters \(\tilde{\mathcal{M}}_{\text{qry}}\) of the query image \(I_{\text{qry}}\) with the same reconstruction model, and then refine it by conditioning on the exemplar set $\mathcal{C}_{\mathcal{M}}$ via the ICL paradigm (as Eq.~\ref{eq:icl}):
\begin{equation}
    \mathcal{M}_{\text{qry}} = \mathcal{F}(\tilde{\mathcal{M}}_{\text{qry}} | \mathcal{C}_{\mathcal{M}}).
\end{equation} 
This formulation establishes the foundation of applying ICL for egocentric 3D hand reconstruction.

\subsection{EgoHandICL Framework}
\label{sec:framework}
\begin{figure}[!t]
    \centering
    \includegraphics[width=\linewidth]{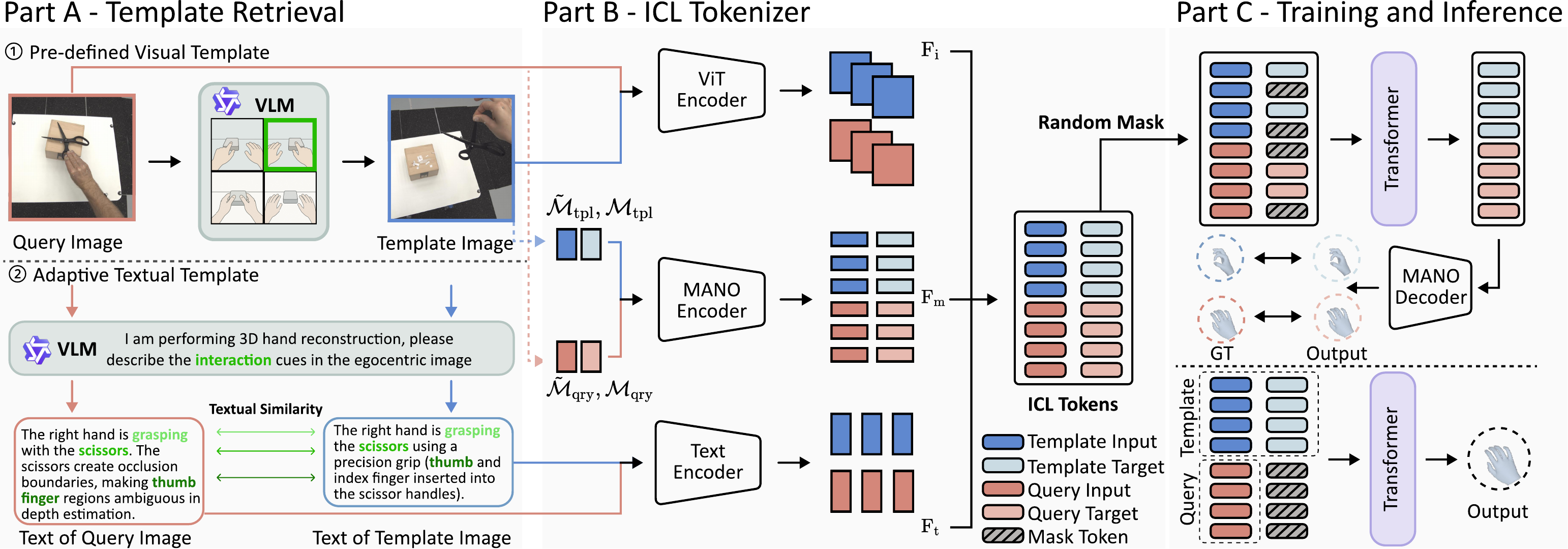}
    \caption{\textbf{Overview of our EgoHandICL framework.} 
        \textbf{Part A:} Given a query image, we retrieve templates via two complementary strategies. 
        \textit{Pre-defined Visual Templates}: a VLM classifies the hand-involvement type and retrieves a template image of the same type. 
        \textit{Adaptive Textual Templates}: we prompt the VLM to generate semantic descriptions, and retrieve a template image given textual similarity. 
        \textbf{Part B:} We encode image tokens \(\text{F}_{\text{i}}\), structural tokens \(\text{F}_{\text{m}}\), and text tokens \(\text{F}_{\text{t}}\), respectively; and then apply cross-attention to tokenize four structured sets of ICL tokens. 
        \textbf{Part C:} We follow a MAE-style design, where the template and query target tokens are partially masked to train the Transformer. 
        In inference, the query target tokens are fully masked for the Transformer's prediction. 
    }
    \label{fig:framework}
\end{figure}

Fig.~\ref{fig:framework} illustrates the overall EgoHandICL framework. 
It is composed of three key components: template retrieval strategies, multimodal context tokenizer, and ICL learning and inference pipelines,
where a VLM is prompted to retrieve the template mainly based on semantic descriptions derived from user-specified prompts regarding interactions, occlusions, and other egocentric cues.

\noindent \textbf{Template Retrieval.} 
As detailed in Sec.~\ref{sec:icl_modeling}, a key step in our framework is retrieving a contextually relevant template image $I_{\text{tpl}}$ for the input query image $I_{\text{qry}}$.
As shown in Fig.~\ref{fig:framework} (Part A), we propose two complementary retrieval strategies that exploit both visual and textual information for contextual adaptation within the ICL paradigm. 

Specifically, we introduce \textit{Pre-defined Visual Templates}, where the VLM classifies each image into one of four pre-defined egocentric hand-involvements: left hand-, right hand-, two hands-, and non hand-involvement types. 
This categorization covers common egocentric hand configurations, allowing us to retrieve visually consistent exemplars from the dataset. 
In addition, hand involvement alone is not always sufficient, as egocentric scenarios often include object interactions, occlusions, missing hands, \emph{etc}. 
To address these issues, we further introduce \textit{Adaptive Textual Templates}, where the VLM retrieves the template mainly based on semantic descriptions derived from user-specified prompts regarding interactions, occlusions, and other egocentric cues.
This strategy enables more flexible and dynamic retrieval of semantically consistent exemplars, providing finer-grained contextual cues in addition to the basic pre-defined visual templates. 
In this work, we retrieve one template image per query image.
Further implementation details are provided in Appendix~\ref{a:egohandicl}.

\noindent \textbf{ICL Tokenizer.} 
Once the template image is retrieved, we construct the ICL tokens by incorporating multimodal context from the template image $I_{\text{tpl}}$ and the query image $I_{\text{qry}}$. 
As shown in Fig.~\ref{fig:framework} (Part B), for both $I_{\text{tpl}}$ and $I_{\text{qry}}$, we estimate their coarse MANO parameters, collect their ground-truth MANO parameters, and feed these four sets of MANO parameters (\emph{i.e.}, $\tilde{\mathcal{M}}_{\text{tpl}}, \tilde{\mathcal{M}}_{\text{qry}}, \mathcal{M}_{\text{tpl}}, \mathcal{M}_{\text{qry}}$) into a MANO encoder, generating the structural tokens \(\text{F}_{\text{m}}\) that preserve 3D hand articulation and shape priors. 
In addition, we encode $I_{\text{tpl}}$ and $I_{\text{qry}}$, respectively, with a pretrained ViT encoder~\citep{potamias2025wilor}, capturing appearance and spatial details and generating the image tokens \(\text{F}_{\text{i}}\). 
Moreover, the VLM-generated semantic descriptions used for retrieval are embedded as the text tokens \(\text{F}_{\text{t}}\) through a text encoder. 
Finally, we apply cross-attention to fuse the multimodal \textit{structural, image, and text tokens}, producing unified ICL tokens for context-aware reasoning. 
Specifically, after processing $I_{\text{tpl}}$ and $I_{\text{qry}}$, our ICL Tokenizer produces four sets of ICL tokens: $T_{\text{tpl}}^{\text{in}}$ (representing ``input'' of ``template''), $T_{\text{tpl}}^{\text{tar}}$ (representing ``target'' of ``template''), $T_{\text{qry}}^{\text{in}}$ (representing ``input'' of ``query''), and $T_{\text{qry}}^{\text{tar}}$ (representing ``target'' of ``query''). Together, these ICL tokens are utilized in the transformer-based reconstruction model as contextual exemplars.

\noindent \textbf{Masked Reconstruction for ICL.} 
Masked Autoencoders (MAE)~\citep{he2022masked} have been widely used in vision tasks, where part of the input tokens is randomly masked and a model is trained to reconstruct the missing ones. 
This architecture has proven effective for learning robust representations and handling incomplete or ambiguous visual signals. 
However, there is a key challenge when applying it to the ICL paradigm: during training, the model has access to the ground truths of both the template and the query samples;
while in inference, the query target is unavailable, since it is exactly what we aim to infer.
To tackle this, prior ICL methods for vision tasks~\citep{bar2022visual, wang2023images, fang2023explore} introduce masking into the target data during training: by partially masking the targets of both template and query samples, the model is trained to predict under incomplete supervision.
As illustrated in Fig.~\ref{fig:framework} (Part C), our EgoHandICL employs a transformer trained for masked reconstruction over ICL tokens. 
During training, we randomly and partially mask the target tokens, \emph{i.e.}, the ICL tokens of $T_{\text{tpl}}^{\text{tar}}$ and $T_{\text{qry}}^{\text{tar}}$, to simulate such incomplete supervision. 
In inference, $T_{\text{qry}}^{\text{tar}}$ is fully masked (unavailable), yet the trained model can decode (reconstruct) the target of query MANO parameters $\mathcal{M}_{\text{qry}}$ from the remaining ICL tokens of contextual exemplars.
In summary, this MAE-driven design offers an effective training-inference architecture for exemplar-conditioned reasoning, enabling in-context hand reconstruction under challenging egocentric settings.

\subsection{Loss Function}
\label{sec:loss}

We follow standard practices in parametric hand reconstruction and employ 
parameter-level ($\mathcal{L}_{mano}$) and vertex-level (\(\mathcal{L}_{V}\)) supervisions to train our EgoHandICL to reconstruct 3D hand meshes:
\begin{equation}
    \mathcal{L}_{mano} = \left\| \Theta - \Theta^{\text{gt}}\right\|^2_2 + \left\| \beta - \beta^{\text{gt}}\right\|^2_2 + \left\| \Phi - \Phi^{\text{gt}}\right\|^2_2, \quad \mathcal{L}_{V} = \left\| V_{3D} - V_{3D}^{\text{gt}} \right\|_1.
\end{equation}
While parameter and vertex supervisions provide essential geometric constraints, they remain insufficient in egocentric settings with lacking enough information only from the input image itself. 
Inspired by perceptual losses~\citep{johnson2016perceptual} used for image restoration, we introduce a hand-specific 3D perceptual loss $\mathcal{L}_{3D}$, which aims to align high-level embeddings of predicted and ground-truth hand meshes. 
This loss enforces semantic consistency under occlusion and ambiguity, leading to reconstructions that are both geometrically accurate and perceptually realistic:
\begin{equation}
    \mathcal{L}_{3D} = \left\| \phi(\mathcal{P}) - \phi(\mathcal{P}^{\text{gt}}) \right\|_2^2,
\end{equation}
where $\mathcal{P}$ is the point cloud formed by the vertices (or joints) of the MANO-generated hand mesh, and $\phi(\cdot)$ is a pretrained 3D feature encoder. 
The overall loss $\mathcal{L}$ can be defined as:
\begin{equation}
    \mathcal{L} = \lambda_{m} \mathcal{L}_{mano} 
               + \lambda_{v} \mathcal{L}_{V} 
               + \lambda_{3D} \mathcal{L}_{3D},
\end{equation}
where $\lambda_{m}$, $\lambda_{v} $, and $\lambda_{3D}$ are empirical weights. 
For datasets without MANO ground truth, \eg, EgoExo4D~\citep{chen2024pcieegohandposesolutionegoexo4dhand}, the loss applies 3D key-joint $J_{3D}$ constraints, weighted by $\lambda_{j}$:
\begin{equation}
    \mathcal{L}_{J} =  \left\| J_{3D} - J_{3D}^{\text{gt}} \right\|_1,
    \quad
    \mathcal{L} = 
                \lambda_{j} \mathcal{L}_{J} 
               + \lambda_{3D} \mathcal{L}_{3D}.
\end{equation}

\subsection{Implementation}
\label{sec:details}
For the retrieval process, we employ Qwen2.5-VL-72B-Instruct~\citep{qwen2.5} as our VLM.
For the ICL tokenizer, we use the same pretrained ViT backbone~\citep{dosovitskiy2020image} as in WiLoR, while the text encoder is Qwen-7B~\citep{bai2023qwentechnicalreport}. 
The MANO encoder and decoder are implemented as MLPs, training together within the EgoHandICL framework. 
We implement the reconstruction model as a lightweight Transformer encoder.
%
We also apply Uni3D-ti~\citep{zhou2023uni3d} as the 3D feature encoder $\phi$ used in the loss $\mathcal{L}_{3D}$. 
All models are trained for 100 epochs with a learning rate of $1 e \!-\! 4$. 
The loss weights are set as \(\lambda_{m}\!=\!0.05\), \(\lambda_{v}\!=\!5.0\), \(\lambda_{j}\!=\!5.0\), \ and   \(\lambda_{3D}\!=\!0.01 \).
The training is conducted on a single RTX 4090 GPU with a batch size of 64 and AdamW optimizer~\citep{loshchilov2018fixing}. 
We use 4 A100 GPUs for data preprocessing and retrieval.

%% file: sec/4_exp.tex
\section{Experiments}
\subsection{Datasets and Evaluation Metrics}
\noindent \textbf{Datasets.} 
We evaluate the EgoHandICL framework on two benchmarks: the ARCTIC dataset~\citep{fan2023arctic} provides high-quality MANO parameter annotations in controlled laboratory environments; and the EgoExo4D dataset~\citep{grauman2024ego} consists of diverse egocentric videos with challenging hand-object interactions captured in realistic, unconstrained scenarios. These two datasets allow us to comprehensively evaluate the 3D hand reconstruction performance of EgoHandICL from the perspectives of hand mesh vertices (ARCTIC, 118.2K training / 16.9K testing samples) and hand skeleton joints (EgoExo4D, 17.3K training / 4.1K testing samples). 

\noindent \textbf{Baselines.} 
We compare the performance of EgoHandICL with state-of-the-art 3D hand reconstruction baselines. For the ARCTIC dataset, we mainly evaluate against 3D hand mesh reconstruction methods, including HaMeR~\citep{pavlakos2024reconstructing}, WiLoR~\citep{potamias2025wilor}, WildHand~\citep{prakash20243d} and HaWoR~\citep{zhang2025hawor}. For the EgoExo4D dataset, we primarily assess against 3D hand joint estimation approaches, including POTTER~\citep{zheng2023potter}, PCIE-EgoHandPose~\citep {chen2024pcieegohandposesolutionegoexo4dhand}, and hand mesh reconstruction methods.

\noindent \textbf{Evaluation Metrics.}
We evaluate egocentric hand reconstruction using both vertex-level and joint-level metrics, under two evaluation settings. In the general setting, metrics are computed for each detected hand, where cases with failed hand detection are excluded from evaluation. For ARCTIC, we report Procrustes-Aligned Mean Per Joint Position Error (P-MPJPE) and Procrustes-Aligned Mean Per Vertex Position Error (P-MPVPE), along with the fraction of vertices within 5mm and 15mm error thresholds (F@5, F@15). For EgoExo4D, we report the Mean Per Joint Position Error (MPJPE) and its Procrustes-aligned variant (P-MPJPE), together with F@10 and F@15 to measure the proportion of accurately reconstructed joints. To further capture the challenges of egocentric views while ensuring fair comparison under consistent detection conditions, we also evaluate under the bimanual setting, where only samples with both hands correctly detected are counted. In addition to P-MPVPE, we compute the Mean Relative Root Position Error (MRRPE) in this setting, to quantify the spatial consistency between the left and right hands~\citep{fan2021learning, fan2023arctic, moon2020interhand2}. All metrics for both vertex-level and joint-level evaluations are reported in millimeters.

\input{table/arctic}
\input{table/egoexo}
\subsection{Experiment Analysis}
\begin{figure}[!t]
    \centering
\includegraphics[width=.97\linewidth]{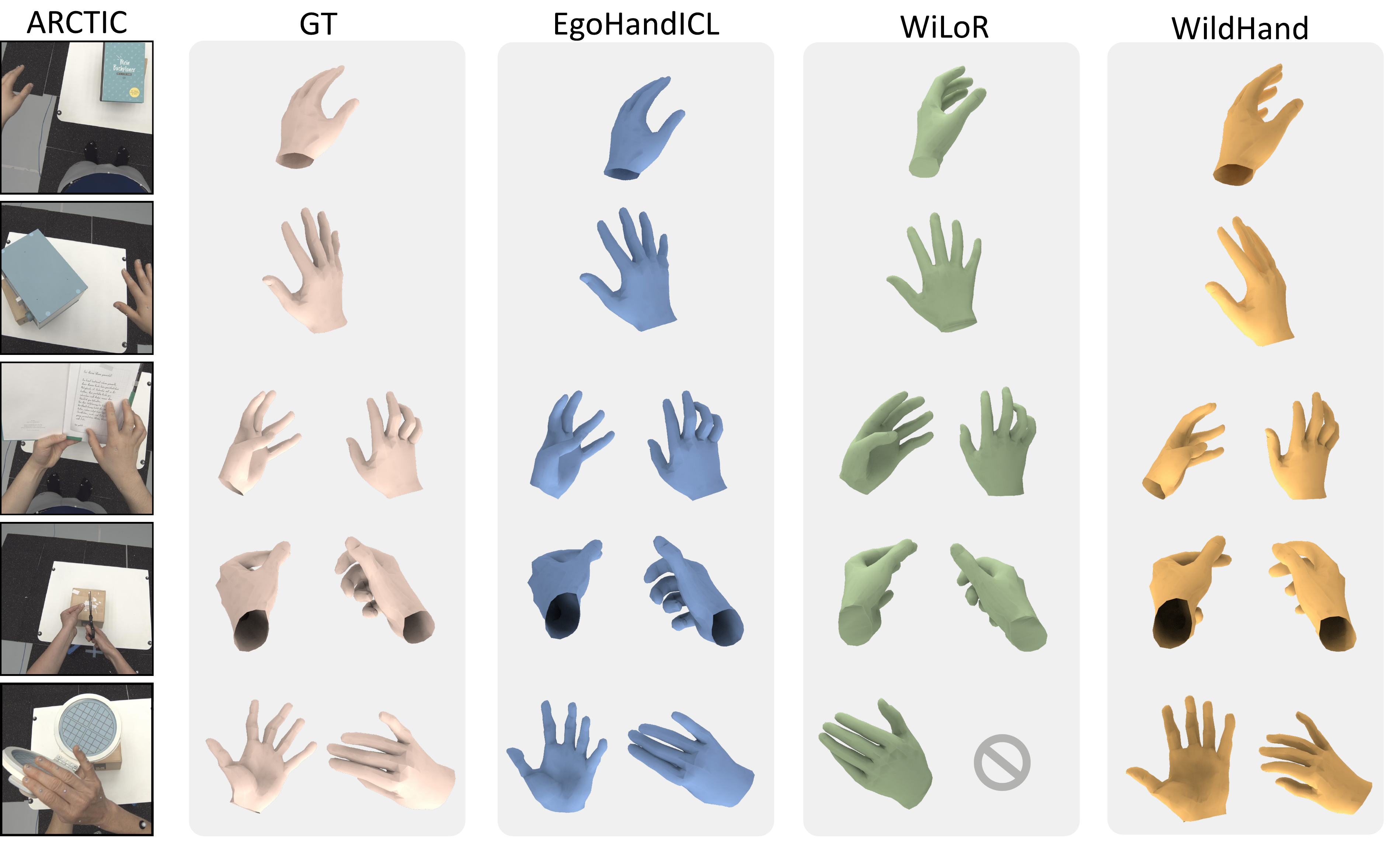}
    \caption{\textbf{Qualitative results on the ARCTIC dataset.}     Note: In the bottom case, where the two hands cross and the left hand is severely occluded, WiLoR~\citep{potamias2025wilor} reconstructs only the right hand but mistakenly identifies it as the left.}
    \label{fig:arctic}
\end{figure}

\begin{figure}[h]
    \centering
    \includegraphics[width=\linewidth]
    {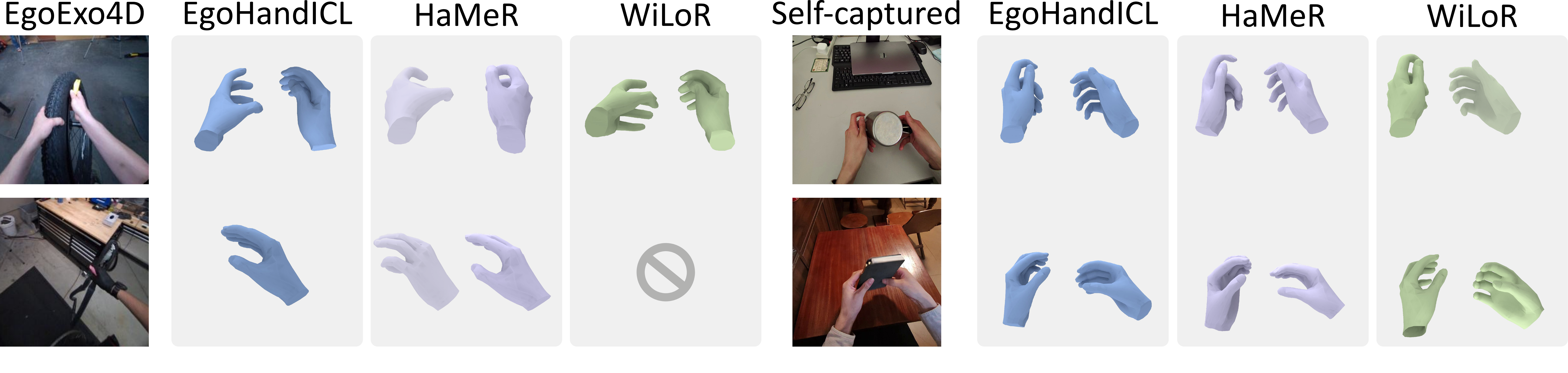}
    \caption{\textbf{Qualitative results on the EgoExo4D dataset (left) and self-captured cases (right)}. Note: In the bottom-left case with a single heavily occluded hand, HaMeR~\citep{pavlakos2024reconstructing} mistakenly reconstructs two hands, whereas WiLoR~\citep{potamias2025wilor} fails to reconstruct any.
 }
    \label{fig:egoexo}
\end{figure}

\noindent \textbf{Experimental Results of 3D Hand Mesh Reconstruction.}
We use the ARCTIC dataset for both joint- and vertex-level evaluations. 
As shown in Tab.~\ref{tab:arctic}, while WiLoR performs well when evaluating only detected hands in the general setting, its accuracy drops in bimanual cases, often failing to recover heavily occluded hands or confusing left/right identities (see the bottom case in Fig.~\ref{fig:arctic}). 
Compared to the second best, EgoHandICL consistently improves PA-MPVPE by 31.1\% and 24.5\% in the general and bimanual settings, respectively, and further reduces MRRPE by 12\%.
This indicates our finding that, with in-context learning, EgoHandICL is particularly effective at resolving egocentric occlusions and estimating spatial relations between the two hands. 
For the EgoExo4D dataset (Tab.~\ref{tab:egoexo}), where MRRPE is substantially higher than in the ARCTIC dataset due to dynamic egocentric viewpoints, EgoHandICL again achieves notable gains with reduced errors, demonstrating strong spatial consistency between the two hands. 
Qualitative comparisons in Figs.~\ref{fig:arctic} and \ref{fig:egoexo} highlight that egocentric reconstruction involves not only single-hand accuracy but also maintaining reasonable relative geometry between the two hands in rapidly changing egocentric views. 
In general, our EgoHandICL consistently improves both visual accuracy and geometric fidelity across scenarios with egocentric occlusions and bimanual interaction, providing a more robust solution for real-world cases. Further discussions are provided in Appendix~\ref{a:exp}.

\input{table/subdataset}
\noindent \textbf{In-context Reasoning Analysis.}
To verify the in-context reasoning abilities of the EgoHandICL framework, as well as the effectiveness of our categorization of pre-defined visual templates, we train variants of EgoHandICL on sub-datasets, each containing only a single hand-involvement type. 
As shown in Tab.~\ref{tab:icl}, each variant performs the best when the evaluation condition matches the hand involvement type of its training templates. 
For example, Proposed-L achieves the lowest errors when testing on the sub-dataset of the left-hand involvement type. 
However, this configuration also introduces side effects: models trained only on a specific sub-dataset (\emph{e.g.}, Proposed-N) generalize poorly to other types of samples (\emph{e.g.}, the Two Hands sub-dataset). 
To this end, our Proposed-Full model, trained on all sub-datasets, achieves the best accuracy across all hand-involvement types, demonstrating the synergistic benefits of in-context learning for stronger generalization and effective adaptation to diverse scenarios.

\input{table/text}

\noindent \textbf{Different Prompts for Retrieval.} 
To assess the role of adaptive textual templates, we compare two prompting strategies for generating retrieval texts via the VLM (Tab.~\ref{tab:text}). 
When using only templates retrieved from the same hand-involvement type without any adaptive textual templates (\textit{W/o.} Prompts), the model achieves relatively stable performance on F@5 and F@15. 
Introducing adaptive textual templates with descriptive prompts (\textit{Des.} Prompts: \textit{e.g.}, ``\textit{I am performing 3D hand reconstruction, please describe the interaction or occlusion details in the egocentric image.}'') reduces reconstruction errors by providing explicit cues about occlusions.
In particular, 
reasoning-style prompts (\textit{Reas.} Prompts: \textit{e.g.}, ``\textit{..., please provide guidance for handling occlusions and complex interactions.}'') further improve accuracy, as they encourage VLM to exploit richer contextual cues with semantic grounding. These results indicate that leveraging VLM’s reasoning capabilities enables more effective and robust in-context learning for hand reconstruction.
Our default setting balances this trade-off by applying the reasoning prompts under heavy occlusion and the description prompts in clearer scenarios.


\subsection{Ablation Analysis}
\begin{center}
\begin{minipage}{\textwidth}
\begin{minipage}[t]{0.48\textwidth}
\makeatletter\def\@captype{table}
\caption{\textbf{Comparison of coarse MANO prediction backbones.} 
Improvement denotes relative gains over the corresponding baseline performance on the ARCTIC dataset in Tab.~\ref{tab:arctic}.}
\label{tab:mano}
\resizebox{0.9\linewidth}{!}{%
\begin{tabular}{lccc}
\specialrule{1.2pt}{0pt}{2pt}
Method & P-MPJPE$\downarrow$ & P-MPVPE$\downarrow$ & F@5$\uparrow$ \\
\toprule
Ours \textit{W}. HaMeR    & 8.3 & 8.1 & 0.052 \\
\textbf{Improvement} &\cellcolor{FIRST} $+16.1\%$ &\cellcolor{FIRST} $+10.4\%$ & \cellcolor{FIRST}$+13.4\%$ \\ 
\midrule
Ours \textit{W}. WildHand  & 5.1 & 4.9 & 0.823 \\
\textbf{Improvement} &\cellcolor{FIRST} $+12.1\%$ &\cellcolor{FIRST} $+12.5\%$ &\cellcolor{FIRST} $+10.3\%$ \\ 
\midrule
Ours \textit{W}. WiLoR & 4.0 & 3.8 & 0.805 \\
\textbf{Improvement} & \cellcolor{FIRST}$+27.3\%$ & \cellcolor{FIRST}$+30.9\%$ &\cellcolor{FIRST} $+7.3\%$ \\ 
\bottomrule
\end{tabular}}
\end{minipage}
\hfill
\begin{minipage}[t]{0.48\textwidth}
\makeatletter\def\@captype{table}
\caption{\textbf{Comparison of different mask ratios for ICL tokens.} Results are tested on the ARCTIC dataset.}
\centering
\resizebox{\linewidth}{!}{%
\begin{tabular}{ccccc}
\specialrule{1.2pt}{0pt}{2pt}
\makecell[c]{Mask\\Ratio} & \makecell[c]{P-MPJPE}$\downarrow$ & \makecell[c]{P-MPVPE}$\downarrow$ & \makecell[c]{F@5$\uparrow$} & \makecell[c]{F@5$\uparrow$}\\
\toprule    
0.4 & 4.3 & 4.2 & 0.761 & 0.995\\
0.5 & 4.1 & 4.1 & 0.788 & 0.996\\
0.6 & 4.1 & 4.1 & 0.791 & 0.997\\
\cellcolor{FIRST}0.7 & \cellcolor{FIRST}4.0 & \cellcolor{FIRST}3.8 & \cellcolor{FIRST}0.801 & \cellcolor{FIRST}0.998\\
0.8 & 4.2 & 4.0 & 0.782 & 0.996\\
\bottomrule
\end{tabular}
\label{tab:mask}}
\end{minipage}
\end{minipage}
\end{center}
\noindent \textbf{Backbone for Coarse MANO Prediction.}
Our EgoHandICL framework is designed to be independent of the choice of coarse MANO prediction backbone. 
To verify this, we use different MANO prediction backbones in our EgoHandICL framework.
As shown in Tab.~\ref{tab:mano}, EgoHandICL consistently and significantly improves over the corresponding baselines. 
These results indicate that the gains are attributed to the in-context learning paradigm itself, demonstrating the generalization and robustness of EgoHandICL regardless of the coarse MANO prediction backbone.

\noindent \textbf{Impact of Mask Ratio for ICL Tokens.}
We evaluate the mask ratio over a broad range (40\%–80\%). As shown in Tab.~\ref{tab:mask}, lower mask ratios moderately degrade performance, while a 70\% mask ratio yields the best results. This suggests that in egocentric hand reconstruction, where occlusions and ambiguous interactions are common, masking a larger portion of ICL tokens encourages the model to exploit stronger contextual cues and employ deeper reasoning to infer hidden structures. The observation is consistent with the key insight of MAE~\citep{he2022masked}, where a higher masking ratio enables the model to learn more informative latent features.

\input{table/3dloss}
\noindent \textbf{3D Perceptual Loss.}
We analyze the role of our introduced hand-specific 3D perceptual loss, which is designed to improve implicit hand representation alignment, complementing geometric parameter supervision. As Tab.~\ref{tab:3dloss} shows, incorporating our 3D perceptual supervision $\mathcal{L}_{3D}$ provides additional benefits over standard geometric regressions $\mathcal{L}_{V}$ and $\mathcal{L}_{mano}$. This indicates that hand-specific 3D perceptual loss $\mathcal{L}_{3D}$ adds complementary 3D cues that are not fully captured by $\mathcal{L}_{V}$ or $\mathcal{L}_{mano}$ alone.

%% file: table/arctic.tex
\begin{table}[!h]
\centering
\caption{\textbf{Quantitative results on the ARCTIC dataset.} We follow the standard evaluation protocol and report both the joint- and vertex-level metrics.}
\resizebox{0.9\linewidth}{!}{%
\begin{tabular}{lcccccc}
\specialrule{1.2pt}{0pt}{2pt}
& \multicolumn{4}{c}{\textit{General Setting}} & \multicolumn{2}{c}{\textit{Bimanual Setting}} \\
\cmidrule(l){2-5} \cmidrule(l){6-7}
\multirow{-2}{*}{Method} & P-MPJPE$\downarrow$ & P-MPVPE$\downarrow$ & F@5$\uparrow$ & F@15$\uparrow$ & P-MPVPE$\downarrow$&MRRPE$\downarrow$ \\
\toprule
HaMeR~\citep{pavlakos2024reconstructing}& 9.9 & 9.6 & 0.046 & 0.911  & 9.9 & 10.1\\
WiLoR~\citep{potamias2025wilor} & \cellcolor{SECOND}5.5 & \cellcolor{SECOND}5.5 & \cellcolor{SECOND}0.524 & \cellcolor{SECOND}0.994 & 5.7 & 9.8 \\
WildHand~\citep{prakash20243d} & 5.8 & 5.6 & 0.746 & 0.928 & \cellcolor{SECOND}4.9 & \cellcolor{SECOND}7.1 \\
HaWoR~\citep{zhang2025hawor} & 6.2 & 6.1 & 0.474 & 0.896 & 6.0 & 8.6\\

\midrule
\textbf{EgoHandICL (ours)} & \cellcolor{FIRST}4.0 & \cellcolor{FIRST}3.8 & \cellcolor{FIRST}0.801 & \cellcolor{FIRST}0.996 & \cellcolor{FIRST}3.7& \cellcolor{FIRST}
6.2\\

\bottomrule
\end{tabular}
}
\label{tab:arctic}
\end{table}

%% file: table/egoexo.tex
\begin{table}[!h]
\centering
\caption{\textbf{Quantitative results on the EgoExo4D dataset.} We follow the standard evaluation protocol and report the joint-level metrics.}
\resizebox{0.9\linewidth}{!}{%
\begin{tabular}{lcccccc}
\specialrule{1.2pt}{0pt}{2pt}
& \multicolumn{4}{c}{\textit{General Setting}} & \multicolumn{2}{c}{\textit{Bimanual Setting}} \\
\cmidrule(l){2-5} \cmidrule(l){6-7}
\multirow{-2}{*}{Method} & MPJPE$\downarrow$ & P-MPJPE$\downarrow$ & F@10$\uparrow$ & F@15$\uparrow$ & P-MPJPE$\downarrow$ & MRRPE$\downarrow$ \\
\toprule
PCIE-EgoHandPose~\citep{chen2024pcieegohandposesolutionegoexo4dhand} & \cellcolor{SECOND}25.5 & \cellcolor{SECOND}8.5 & \cellcolor{SECOND}0.544 & \cellcolor{SECOND}0.910 & \cellcolor{SECOND}8.2 & \cellcolor{SECOND}130.9\\
Potter~\citep{zheng2023potter} & 28.9 & 11.1 & 0.491 & \cellcolor{SECOND}0.910 & 10.3 & 148.9\\
HaMeR~\citep{pavlakos2024reconstructing} & 30.1 & 11.2 & 0.453 & 0.883 & 10.6 & 361.2\\
WiLoR~\citep{potamias2025wilor} &31.1 & 12.5 & 0.528 & 0.905 & 11.0 & 378.0 \\
\midrule
\textbf{EgoHandICL (ours)} & \cellcolor{FIRST}21.1 & \cellcolor{FIRST}7.7 & \cellcolor{FIRST}0.789 & \cellcolor{FIRST}0.935 & \cellcolor{FIRST}7.5 & \cellcolor{FIRST}110.9\\
\bottomrule
\end{tabular}
}
\label{tab:egoexo}

\end{table}

%% file: table/subdataset.tex
\begin{table}[h]
\caption{\textbf{In-context reasoning analysis across different hand-involvement types.} L, R, T, and N denote training on the left-hand, right-hand, two-hand, and non-hand involvement type sub-dataset, respectively. Results are tested on the ARCTIC under these four sub-dataset divisions.}
\centering
\resizebox{\linewidth}{!}{%
\begin{tabular}{lcccccccc}
\specialrule{1.2pt}{5pt}{2pt}
Type & \multicolumn{2}{c}{Left Hand} & \multicolumn{2}{c}{Right Hand} & \multicolumn{2}{c}{Two Hands} & \multicolumn{2}{c}{Non Hand} \\
\cmidrule(l){1-1} \cmidrule(l){2-3} \cmidrule(l){4-5} \cmidrule(l){6-7} \cmidrule(l){8-9}
Method & P-MPJPE$\downarrow$ & P-MPVPE$\downarrow$ & P-MPJPE$\downarrow$ & P-MPVPE$\downarrow$ & P-MPJPE$\downarrow$ & P-MPVPE$\downarrow$ & P-MPJPE$\downarrow$ & P-MPVPE$\downarrow$ \\
\toprule
Proposed - L & \cellcolor{SECOND}4.6 & \cellcolor{SECOND}4.4 
& 4.6 & 4.5
& 5.1 & 4.8
& 5.2 & 5.1  \\
Proposed - R & 4.8 & 4.5
& \cellcolor{SECOND}4.1 & \cellcolor{SECOND}4.4
& 5.0 & 4.8
& 5.2 & 5.2  \\
Proposed - T & 4.7 & 4.6 & 
4.9 & 4.4 & 
\cellcolor{SECOND}4.3 & \cellcolor{SECOND}4.2 
& \cellcolor{SECOND}5.0 & 5.1  \\
Proposed - N & 4.9 & 4.7 & 
5.3 & 5.1 & 
5.5 & 5.1 & 
\cellcolor{SECOND}5.0 & \cellcolor{SECOND}5.0  \\
\midrule
\textbf{Proposed - Full} & \cellcolor{FIRST} 4.5 & \cellcolor{FIRST}4.3 & \cellcolor{FIRST}4.0 & \cellcolor{FIRST}3.9 & \cellcolor{FIRST}3.9 & \cellcolor{FIRST}3.7 & \cellcolor{FIRST}4.7 & \cellcolor{FIRST}4.5  \\
\bottomrule
\end{tabular}
}
\label{tab:icl}
\end{table}

%% file: table/text.tex

\begin{wrapfigure}{r}{0.5\linewidth} 
\centering
\captionof{table}{\textbf{Comparison of different prompts for adaptive textual templates retrieval.} Results are tested on the ARCTIC dataset.}
\resizebox{\linewidth}{!}{%
\begin{tabular}{lcccc}
\specialrule{1.2pt}{0pt}{2pt}
Method & P-MPJPE$\downarrow$ & P-MPVPE$\downarrow$ & F@5$\uparrow$ & F@15$\uparrow$\\
\toprule
\textit{W/o.} Prompts & 4.3 & 3.9 & \cellcolor{FIRST}0.838 & \cellcolor{FIRST}0.996\\
\textit{Des.} Prompts & 4.2 & \cellcolor{FIRST}3.7 & 0.781 & 0.978\\
\textit{Reas.} Prompts & \cellcolor{FIRST}3.9 & \cellcolor{FIRST}3.7 & 0.766 & 0.975\\
\bottomrule
\end{tabular}}
\label{tab:text}
\end{wrapfigure}

%% file: table/3dloss.tex
\begin{wraptable}{r}{0.48\linewidth} 
\caption{\textbf{Comparison of different loss items for the EgoHandICL training.} Results are tested on the ARCTIC dataset.}
\centering
\resizebox{\linewidth}{!}{%
\begin{tabular}{lccc}
\specialrule{1.2pt}{0pt}{2pt}
Loss function & P-MPVPE$\downarrow$ & F@5$\uparrow$ & F@15$\uparrow$\\
\toprule
$\mathcal{L}_{V}$ & 4.7 & 0.6 & 0.982 \\
$\mathcal{L}_{V} + \mathcal{L}_{mano}$ & 4.3 & 0.6 & 0.994 \\
$\mathcal{L}_{V} + \mathcal{L}_{3D}$ & 3.9 & 0.7 & 0.998 \\
$\mathcal{L}_{V} + \mathcal{L}_{3D} + \mathcal{L}_{mano}$ & \cellcolor{FIRST}3.8 & \cellcolor{FIRST}0.8 & \cellcolor{FIRST}0.998 \\
\bottomrule
\end{tabular}}
\label{tab:3dloss}
\end{wraptable}

%% file: sec/5_application.tex
\section{Exploring EgoVLMs with EgoHandICL}
\input{table/egovlm_r}

\begin{figure}[htbp]
    \centering
    \includegraphics[width=0.82\linewidth]{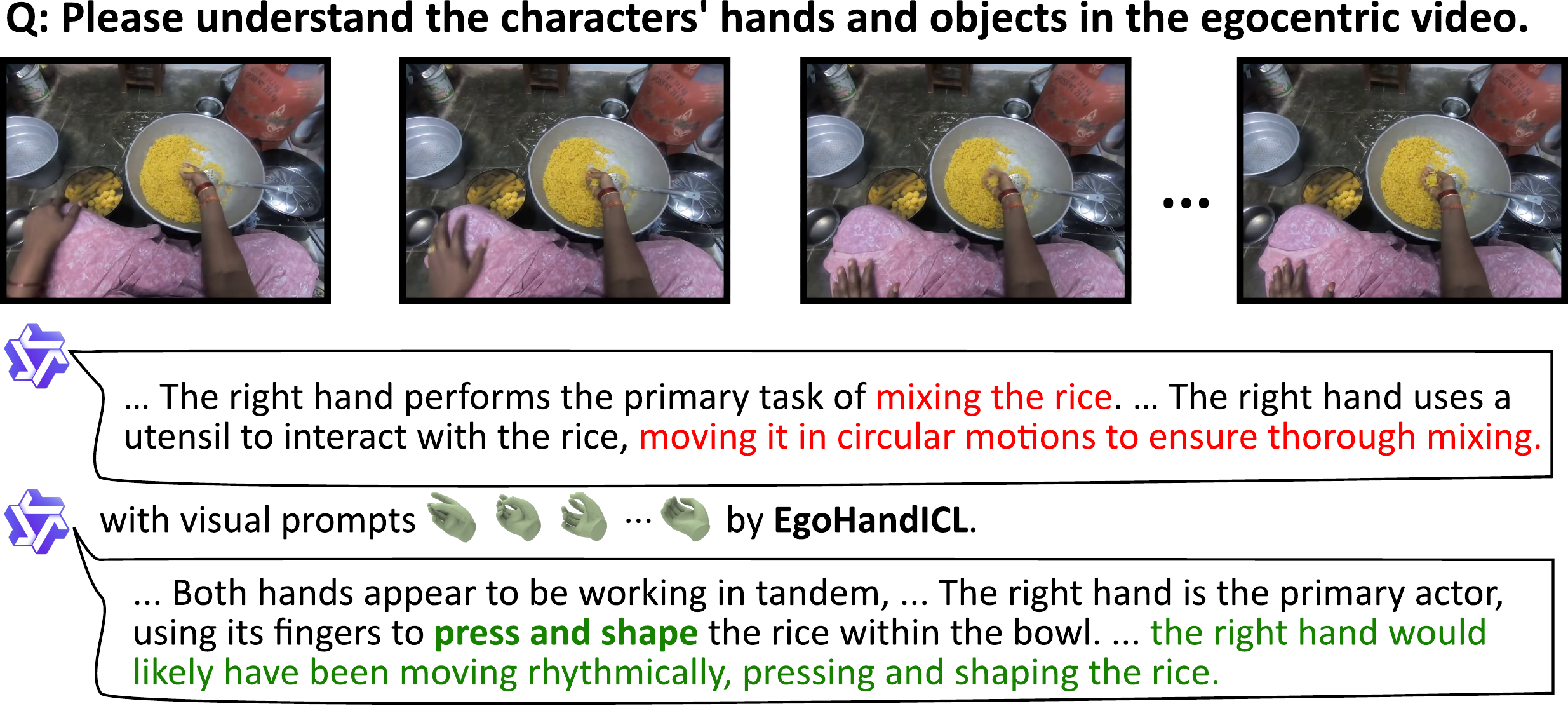}

    \caption{\textbf{EgoVLM's hand–object interaction reasoning with and without EgoHandICL.} By incorporating our hand reconstructions as visual prompts, hand-related actions in egocentric videos can be recognized reliably with finer details.}

    \label{fig:egovlm}
\end{figure}
Recent work has investigated egocentric video-language models (EgoVLMs) for hand-object interaction understanding. To explore whether EgoHandICL can facilitate this task, we evaluate on the EgoHOIBench~\citep{xu2025do} dataset with different EgoVLMs. 
As shown in Tab.~\ref{tab:egovlm}, EgoGPT~\citep{Yang_2025_CVPR}, which is built on LLaVA-OneVision~\citep{li2024llava} and finetuned on a wide range of egocentric data, underperforms the LLaVA-OneVision base model. This suggests that excessive finetuning on egocentric data may bias EgoVLMs toward event-level understanding at the cost of fine-grained hand–object interaction reasoning. To mitigate this issue, we feed EgoHandICL’s reconstruction outputs to the EgoVLM as additional visual prompts, consistently improving their hand-object interactions reasoning capabilities (an example shown in Fig.~\ref{fig:egovlm}). This exploration demonstrates the practical value of EgoHandICL for downstream applications.

%% file: table/egovlm_r.tex
\begin{wrapfigure}{r}{0.48\linewidth}
    \begin{minipage}{\linewidth}
        \vspace{-10pt}
        \captionof{table}{\textbf{Comparison of EgoVLMs on hand-object interaction reasoning.}}
        \vspace{-10pt}
        \centering
        \resizebox{\linewidth}{!}{%
        \begin{tabular}{lcccc}
        \specialrule{1.2pt}{0pt}{2pt}
        Model & avg.$\uparrow$ & verb.$\uparrow$ & none.$\uparrow$ & action$\uparrow$ \\
        \toprule    
        LLaVA-OneVision & 0.75 & 0.68 & 0.82 & 0.58 \\
        + Proposed & \cellcolor{FIRST}0.78 &\cellcolor{FIRST} 0.71 &\cellcolor{FIRST} 0.84 &\cellcolor{FIRST} 0.61\\
        \midrule
        EgoGPT & 0.71 & 0.66 & 0.76 & 0.46 \\
        + Proposed & \cellcolor{FIRST}0.76 & \cellcolor{FIRST}0.70&\cellcolor{FIRST} 0.90 &\cellcolor{FIRST} 0.61\\
        \midrule
        Qwen2.5-VL-7B-Instruct & 0.82 & 0.74 & 0.90 & 0.64\\
        + Proposed &\cellcolor{FIRST} 0.85 & 0.74&\cellcolor{FIRST} 0.91 &\cellcolor{FIRST} 0.69\\
        \bottomrule
        \end{tabular}}
        \label{tab:egovlm}
    \end{minipage}
\end{wrapfigure}

%% file: sec/6_conclusion.tex
\section{Conclusion and Future Work}
We present EgoHandICL, a novel framework for egocentric 3D hand reconstruction with in-context learning. Our method features three main components: (1) VLM-guided retrieval of exemplar templates; (2) an ICL tokenizer synergizing multimodal context; and (3) an MAE-based architecture trained with specialized 3D geometric and perceptual constraints. Extensive experiments on benchmarks and real-world applications validate the effectiveness of our approach. While effective, the computational cost of the VLM-based retrieval presents a limitation for real-time deployment. In the future, we plan to refine the training pipeline and generalize our EgoHandICL framework to broader egocentric settings, such as 3D hand pose tracking and hand–object reconstruction.

%% file: sec/appendix.tex
\clearpage
\appendix
\numberwithin{table}{section}
\numberwithin{figure}{section}
\setcounter{page}{1}
\section*{Appendix}
\addcontentsline{toc}{section}{Appendix}

\section{Discussions}
\subsection{Limitations}
While EgoHandICL achieves strong performance on egocentric 3D hand reconstruction, it has several limitations. 
First, the reliance on VLM-based retrieval introduces a non-trivial computational overhead, which limits real-time deployment on resource-constrained devices. 
Second, the framework depends on high-quality template databases, and retrieval performance may degrade if templates lack diversity. 
Third, egocentric datasets with complete annotations remain limited. For example, the widely used EgoExo4D dataset provides only keypoint-level ground truth for hand pose estimation, lacking full MANO parameter supervision as available in ARCTIC. 
Relying solely on keypoint annotations restricts model generalization to complex real-world scenarios. 
Moreover, most current evaluations are performed with ground-truth bounding boxes, while robust hand detection itself is a major challenge in egocentric views. 
The absence of a fair end-to-end benchmark that jointly evaluates detection and reconstruction limits progress toward more realistic deployments.

\subsection{Future Work}
Building on the identified limitations, several promising directions emerge for future work. 
First, to mitigate the computational overhead of VLM-based retrieval, we plan to explore more efficient retrieval mechanisms, such as lightweight VLMs, approximate nearest-neighbor search, or retrieval-free strategies that embed contextual exemplars directly into the learning pipeline.
Second, addressing the limited availability of fully annotated egocentric datasets, we emphasize the importance of constructing comprehensive 3D egocentric benchmarks with consistent annotations of hand pose, object pose, and body pose. 
Such datasets would enable more holistic evaluation of hand–object–body interactions and facilitate robust generalization in real-world scenarios. 
Third, to move beyond evaluations that rely on ground-truth bounding boxes, we advocate for the construction of fair end-to-end benchmarks that jointly assess detection and reconstruction in egocentric settings, better reflecting real deployment challenges. 
Finally, we will explore temporal extensions of EgoHandICL for continuous 3D hand tracking and broaden its scope to encompass richer egocentric tasks, such as joint hand–object reconstruction and gaze-conditioned interaction modeling, with a particular emphasis on lightweight designs suitable for AR/VR applications.

\subsection{Social Impact}
\label{a.3}
Egocentric 3D hand reconstruction has promising applications in extended reality, assistive technologies, and robotics. However, it also introduces potential risks. 
First, continuous capture of egocentric data may raise privacy concerns, particularly regarding bystanders or sensitive environments. 
Second, the reconstructed hand data could be misused in surveillance or unauthorized tracking applications. 
To mitigate these risks, we emphasize that \textbf{all datasets used in this work comply with their official licenses and community standards, and we strictly adhere to ethical guidelines throughout data usage and research practices.}

\subsection{LLM Usage} Large language models (LLMs) are used to moderately polish the paper writing. Specifically, LLMs are employed for improving grammar and formatting consistency of LaTeX content.

\section{More Details of EgoHandICL}
\label{a:egohandicl}
\subsection{Pre-defined Visual Templates}
We first introduce the construction of \textit{Pre-defined Visual Templates}, which categorize egocentric images into four canonical hand-involvement types: left-hand, right-hand, two-hand, and non-hand involvement. 

\paragraph{Prompt Design.} 
To enable a VLM to automatically categorize each image, we design explicit textual prompts describing the four hand-involvement types. Our prompt is:  
\begin{lstlisting}[language=Python]
SYSTEM_PROMPT = """
You are an image understanding agent. Your task is to analyze a first-person perspective image and classify the interaction status of the left and right hands.

Classification rules:
- 0: Only the left hand is interacting with an object
- 1: Only the right hand is interacting with an object
- 2: Both hands are interacting with an object
- 3: Neither hand is interacting with any object

Important notes:
- Occlusion caused by objects must be considered in determining whether a hand is interacting.
- Use your best reasoning based on the visual content to make this decision.

Your response must be a single number: one of [0, 1, 2, -1]. Do not include any explanation or additional text.
"""

USER_PROMPT = """
Please analyze the first-person perspective image and determine the interaction status of the hands.

Return only the correct label number based on the following:
- 0: Only the left hand is interacting
- 1: Only the right hand is interacting
- 2: Both hands are interacting
- 3: Neither hand is interacting

"""
    
\end{lstlisting}

This prompt is fixed across both training and inference to ensure consistency.

\paragraph{Data Distribution.} 
Based on the above categorization, we partition the dataset into the four hand-involvement subsets. The overall distribution of samples across categories is summarized in Tab. ~\ref{tab:template_distribution}. This categorization allows us to balance template retrieval across common egocentric scenarios.

\begin{table}[h]
\centering
\caption{\textbf{Data distribution of pre-defined visual templates.} 
Each hand-involvement type is split into training, validation, and testing sets.}
\vspace{4pt}
\resizebox{0.8\linewidth}{!}{
\begin{tabular}{c|r r r|r}
\specialrule{1.2pt}{0pt}{2pt}
\textbf{Hand Involvement Type} & \textbf{Train} & \textbf{Val} & \textbf{Test} & \textbf{Total} \\
\midrule
Left Hand & 7,991  & 2,283  & 1,143  & 11,417 \\
Right Hand & 14,334 & 4,095  & 2,049  & 20,478 \\
Both Hand & 92,917 & 26,547 & 13,275 & 132,739 \\
None Hand & 2,996  & 856    & 428    & 4,280 \\
\midrule
\textbf{Total} & \textbf{118,238} & \textbf{33,781} & \textbf{16,895} & \textbf{168,914} \\
\specialrule{1.2pt}{0pt}{2pt}
\end{tabular}}
\label{tab:template_distribution}
\end{table}

\paragraph{Training Strategy.} 
During training-time data preprocessing, we perform the template-retrieval procedure three times for every candidate training sample. This repetition is used to verify the correctness of hand-involvement classification and ensure that each exemplar is assigned to the proper category. Only samples with consistent predictions across all three runs are kept in the exemplar pool.
After this offline verification, the training process itself uses random sampling within each validated subset to construct the exemplar set for each query image at every iteration. This stochastic sampling maintains exemplar diversity and mitigates overfitting.

\paragraph{Inference Protocol.} 
At inference time, we adopt the same VLM prompt to categorize the query image and retrieve templates from the training set accordingly.

\subsection{Adaptive Textual Templates}

\paragraph{Prompt Design.}
We design two families of prompts with different semantic emphases:
\begin{itemize}[noitemsep,leftmargin=*]
    \item \textbf{Description-style prompts (Des. Prompts):} These provide explicit cues about the egocentric scene, encouraging the VLM to describe visible details. Our Prompts are:
    \begin{lstlisting}
Please analyze the egocentric image and return one concise sentence describing the visible hands and their interactions with surrounding objects for 3D hand reconstruction. The description should specify:

1. Which hand(s) are present (left hand, right hand, or both).
2. The type of interaction (e.g., grasping, holding, touching, pointing, resting).
3. The object involved, if any (e.g., a cup, phone, keyboard).
4. If no hand is clearly visible, explicitly state "no hand involvement."

Examples:
- "Left hand grasping a cup."
- "Right hand pointing at a phone."
- "Both hands holding a book."
- "No hand involvement."

    \end{lstlisting}
    The prompts is especially effective when the hand is clearly visible, helping the VLM retrieve templates with matching appearance and interaction patterns.
    \item \textbf{Reasoning-style prompts (Reas. Prompts):} These encourage the VLM to infer strategies for handling challenging cases. Our Prompts are:
    \begin{lstlisting}
SYSTEM_PROMPT = """
You are a reasoning agent specialized in egocentric hand-object interaction understanding. Your task is to analyze images captured from a first-person perspective and generate a caption that describes all hand and object interaction details relevant to 3D hand reconstruction.

Given a single input image, output one concise sentence that specifies which hand(s) (left/right/both) are visible, their pose, and their interaction with any objects. Be precise and descriptive about the hand-object interaction without adding explanations or speculations.
"""

USER_PROMPT = """
Please analyze the image and return one sentence describing the hands and their interactions with objects for hand reconstruction (e.g., left hand grasping a cup, right hand resting on the table).
"""
    \end{lstlisting}
    The prompt is mostly useful for heavily occluded or ambiguous views, where direct visual cues are insufficient and deeper contextual reasoning is required.
\end{itemize}

\subsection{Examples.}
\begin{figure}
    \centering
    \includegraphics[width=\linewidth]{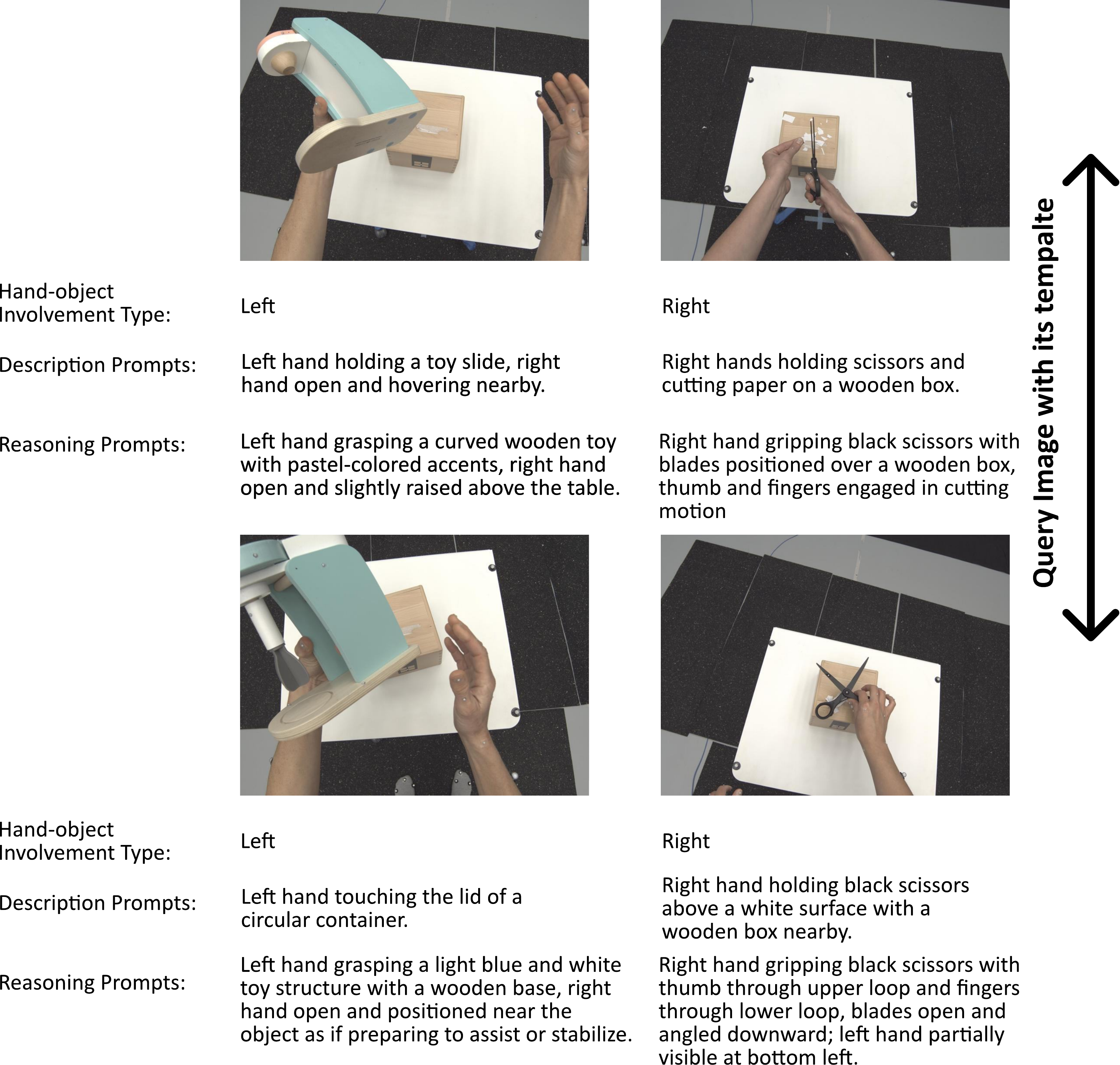}
    \caption{\textbf{Query-Template Example 1.}}
    \label{fig:pair1}
\end{figure}

\begin{figure}
    \centering
    \includegraphics[width=\linewidth]{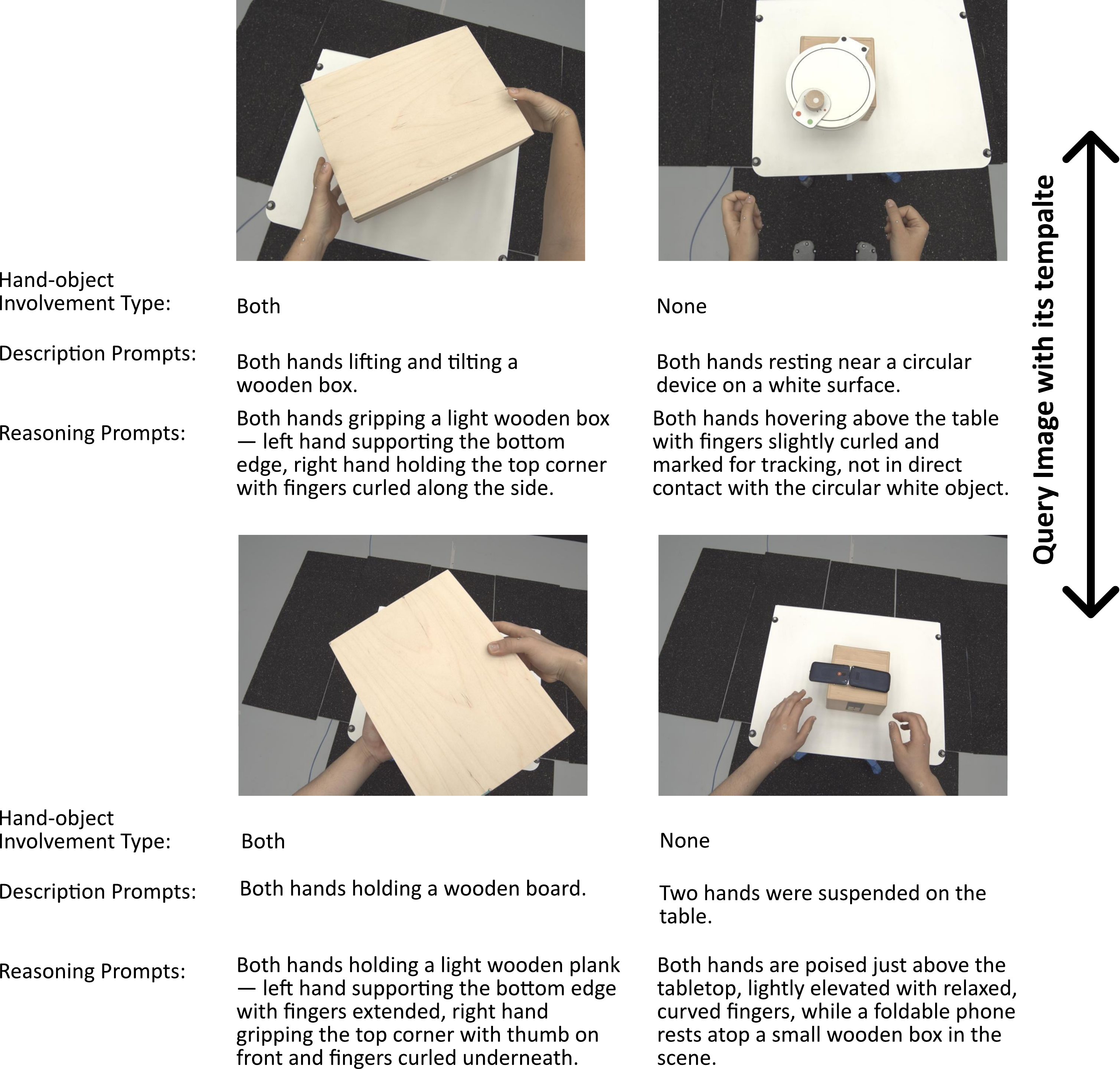}
    \caption{\textbf{Query-Template Example 2.}}
    \label{fig:pair2}
\end{figure}

In Fig.~\ref{fig:pair1} and Fig.~\ref{fig:pair2}, we illustrate examples of the two template retrieval strategies. 
During both training and inference, the retrieved images and the query image can serve as interchangeable 
query–template pairs, enabling flexible construction of in-context demonstrations.


\section{More Details of Experiments}
\label{a:exp}

\paragraph{Evaluation Settings.} 
All baselines are fine-tuned on the same training splits, without incorporating any external data, ensuring a fully fair comparison. In the \textit{General setting}, evaluation is performed on all hands detected within each image. However, egocentric scenarios often involve challenging cases such as partially invisible hands, severe self-occlusions, or multiple interacting hands from different persons. Under such conditions, different baseline models occasionally fail to detect all hands, leading to a small proportion of missing detections. Across our benchmarks, we observe that the proportion of such missing cases remains below 5\%, and therefore the reported results in the general setting are still representative of overall performance. 

\paragraph{Bimanual setting.} 
To provide a stricter and fairer comparison, we further adopt the \textit{Bimanual setting}, where only samples with both hands correctly detected by each evaluated method are considered. This ensures that reconstruction accuracy is compared under consistent detection conditions across models, thereby eliminating potential biases caused by unequal detection performance. The resulting bimanual test set contains 12{,}041 and 3{,}802 samples on ARCTIC and EgoExo4D respectively, providing a robust basis for evaluating hand-to-hand spatial consistency and overall reconstruction reliability in dual-hand interactions.
\paragraph{Performance across hand-involvement categories.} 
To better understand the behavior of different models under varying egocentric conditions, we further evaluate them separately on the four pre-defined hand-involvement types: left-hand, right-hand, two-hand, and no-hand. 
This breakdown allows us to quantify how well each model generalizes to different interaction settings, and to analyze whether certain models are more prone to performance degradation in specific categories.

\begin{table}[h]
\caption{\textbf{Performance of Model Zoos across different hand-involvement types.}}
\vspace{-10pt}
\centering
\resizebox{\linewidth}{!}{%
\begin{tabular}{lcccccccc}
\specialrule{1.2pt}{5pt}{2pt}
Type & \multicolumn{2}{c}{Left Hand} & \multicolumn{2}{c}{Right Hand} & \multicolumn{2}{c}{Two Hands} & \multicolumn{2}{c}{Non Hand} \\
\cmidrule(l){1-1} \cmidrule(l){2-3} \cmidrule(l){4-5} \cmidrule(l){6-7} \cmidrule(l){8-9}
Method & P-MPJPE$\downarrow$ & P-MPVPE$\downarrow$ & P-MPJPE$\downarrow$ & P-MPVPE$\downarrow$ & P-MPJPE$\downarrow$ & P-MPVPE$\downarrow$ & P-MPJPE$\downarrow$ & P-MPVPE$\downarrow$ \\
\toprule
HaMeR~\citep{pavlakos2024reconstructing} & 10.5 & 10.2
& 10.1 & 9.7
& 10.0 & 9.6
& 7.6 & 7.3  \\
WiLoR~\citep{potamias2025wilor} & 5.6 & 5.5
& 5.2 & 5.1
& 5.5 & 5.4
& 5.1 & 5.0  \\
WildHand~\citep{prakash20243d} & 5.9 & 5.8
& 5.6 & 5.4
& 5.3 & 5.1
& 5.2 & 5.0  \\
HaWoR~\citep{zhang2025hawor} & 6.7 & 6.1
& 6.2 & 5.9
& 5.3 & 5.0
& 4.8 & 4.8  \\
\midrule
Average & 7.2 & 6.9
& 6.8 & 6.5
& 6.5 & 6.3
& 5.7 & 5.5  \\
\bottomrule
\end{tabular}
}
\label{tab:modelzoo}
\vspace{-10pt}
\end{table}

As shown in Table~\ref{tab:modelzoo}, the reconstruction errors vary across the four hand-involvement types. 
We observe that left-hand cases exhibit slightly higher errors (7.2/6.9) compared to right-hand cases (6.8/6.5). 
This asymmetry is consistent with prior observations that egocentric datasets are often right-hand dominant (e.g., ARCTIC~\citep{fan2023arctic}, EgoExo4D~\citep{chen2024pcieegohandposesolutionegoexo4dhand}), leading to stronger representations 
for the right hand. Interestingly, two-hand scenarios achieve lower errors (6.5/6.3) than single-hand cases, 
suggesting that the presence of both hands provides additional geometric constraints that help the model resolve occlusions and perspective ambiguities. 
This trend contrasts with conventional baselines such as HaMeR~\citep{pavlakos2024reconstructing} and WiLoR~\citep{potamias2025wilor}, which typically degrade in bimanual settings due to severe inter-hand occlusions. 
Finally, the lowest errors are obtained in the no-hand category (5.7/5.5), demonstrating the model’s robustness in avoiding false positives when no hand is present. 

\section{Retrieval Quality and OOD Analysis.}

\begin{figure}[htbp]
    \centering
    \includegraphics[width=1\linewidth]{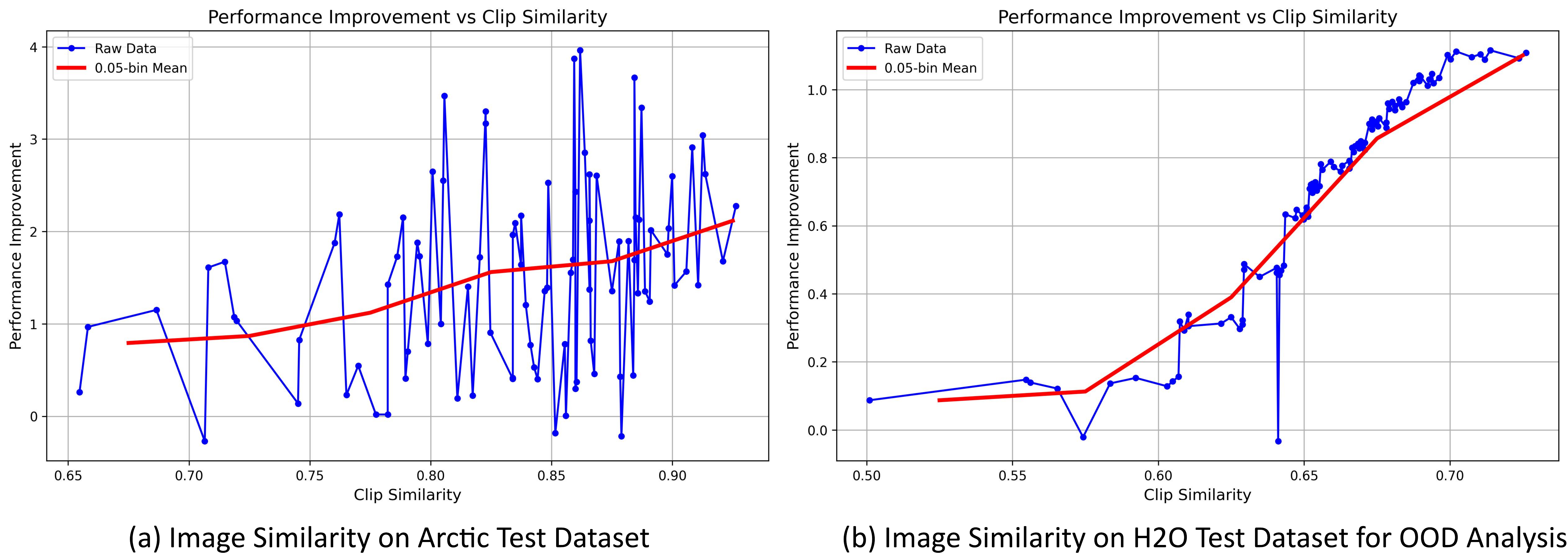}
    \caption{\textbf{Retrieval Quality and OOD Analysis.}}
    \label{fig:ood}
\end{figure}

\subsection{Retrieval Quality Analysis}
On the ARCTIC dataset, we assess the quality of the retrieved templates by computing the \textit{cosine similarity} between the CLIP~\citep{pmlr-v139-radford21a} image features of the query and its exemplar. We further correlate this similarity with the performance gain over the strongest baseline WiLoR (in P-MPVPE). As shown in Figure~\ref{fig:ood} (a), over 90\% of retrieved exemplars achieve a similarity above 0.7, indicating that our retrieval stage provides high-quality contextual matches. Moreover, higher similarity consistently leads to larger improvements over WiLoR, confirming that EgoHandICL effectively leverages high-quality exemplars for contextual refinement of 3D hand reconstruction.

\subsection{Out-of-Distribution (OOD) Analysis}
We further analyze EgoHandICL under OOD conditions on the H2O dataset~\citep{Kwon_2021_ICCV}, where retrieved templates are less aligned with the query image due to distribution shift. As shown in  Figure~\ref{fig:ood} (b), the cosine similarity between the query and its retrieved exemplar is predominantly concentrated in the range [0.6, 0.7), reflecting the difficulty of retrieval in OOD scenarios. Correspondingly, the absolute improvement over WiLoR is smaller than in the in-distribution case Figure~\ref{fig:ood} (a), but EgoHandICL still consistently outperforms WiLoR across the observed similarity range, demonstrating robust generalization of our exemplar-driven ICL framework even when template quality is degraded.

\begin{figure}[h]
    \centering
    \includegraphics[width=1\linewidth]{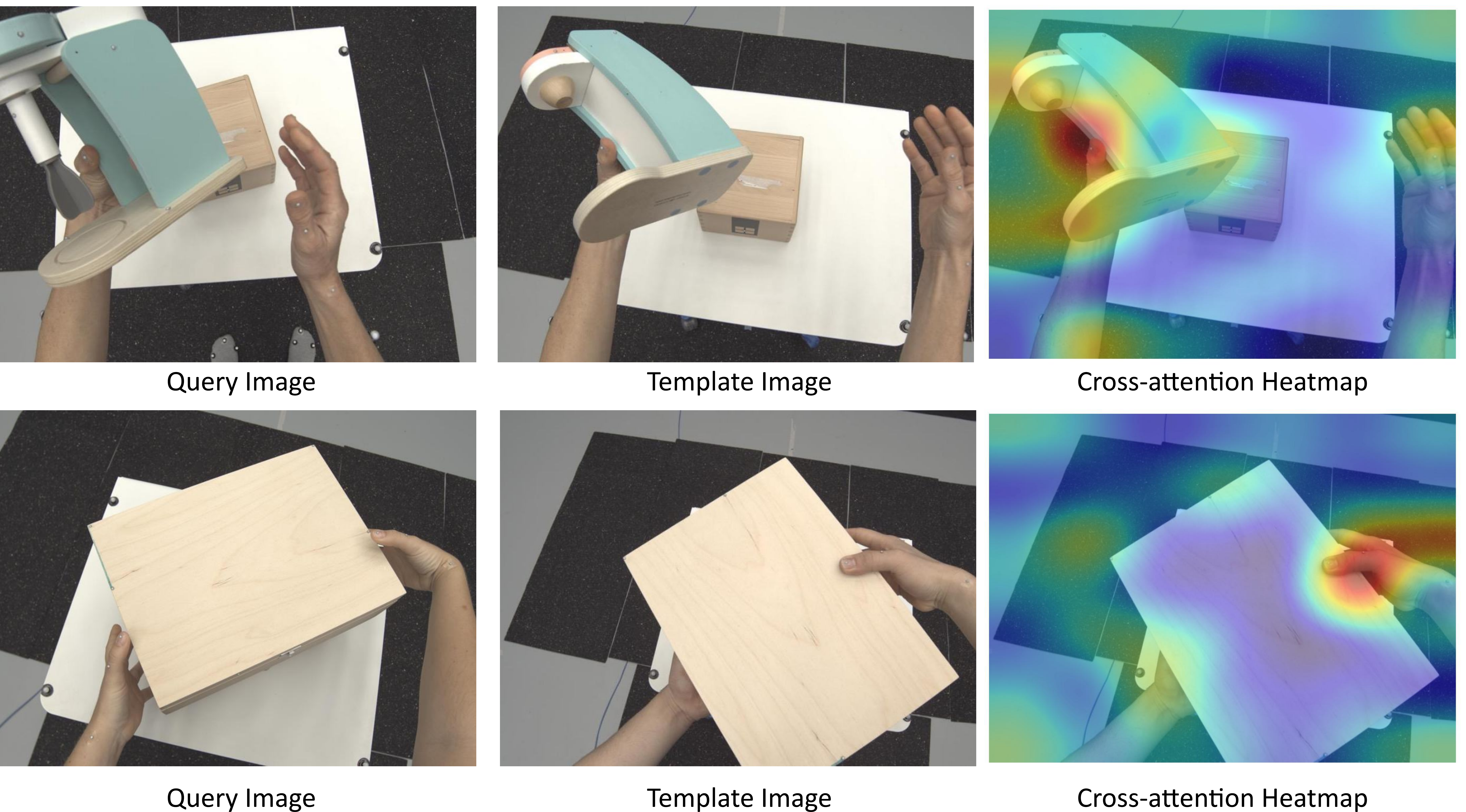}
    \caption{\textbf{Attention-weight Inspection Visualization.}}
    \label{fig:attention}
\end{figure}
\section{Attention-Based Qualitative Visualization}
To further examine how EgoHandICL improves hand reconstruction under occlusion, we present new qualitative visualizations in Figure~\ref{fig:attention} that explicitly illustrate how the model uses the retrieved exemplar to resolve occluded regions in the query.  We perform attention-weight inspection between exemplar tokens and query tokens, visualizing the cross-attention matrix of our ICL Transformer.  The resulting heatmaps consistently show that tokens originating from occluded regions in the query place high-magnitude attention on structurally informative regions of the exemplar, demonstrating that the exemplar provides visual priors the model actively consults during inference.  Overall, these visualizations directly confirm that EgoHandICL performs genuine contextual reasoning, enabling the model to recover accurate 3D hand reconstruction under occlusion.

\section{Failure Case}
\begin{figure}[htbp]
    \centering
    \includegraphics[width=\linewidth]{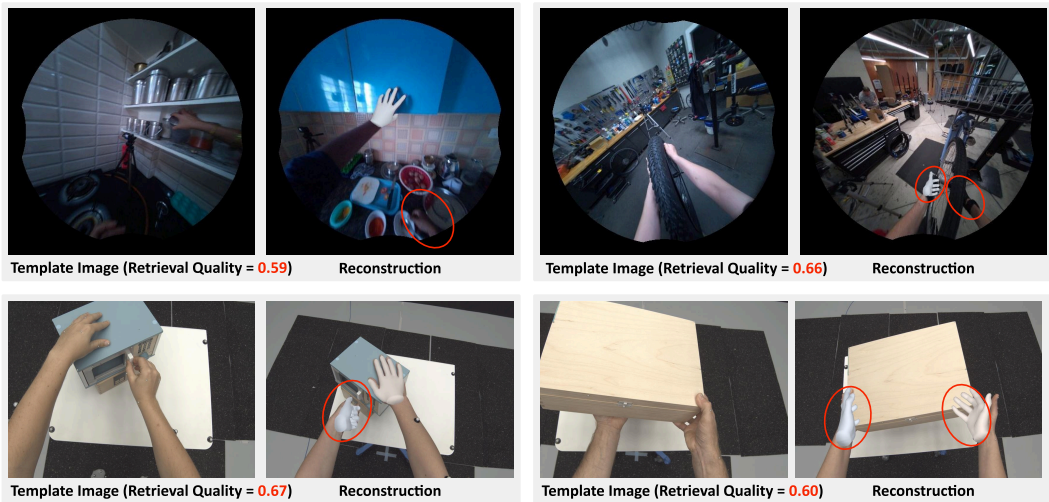}
    \caption{\textbf{Failure Cases.} Failure cases under low-quality retrieval.}
    \label{fig:failure}
\end{figure}
As shown in Figure~\ref{fig:failure}, EgoHandICL may fail when the retrieved exemplar is of low quality. Such low-quality retrieval typically arises under challenging visual conditions (e.g., severe occlusion, motion blur, or non-typical hand–object configurations), where the visual evidence becomes insufficient for accurate matching. In these scenarios, the retrieved exemplar offers weak contextual cues, leading to degraded reconstruction quality. Also, these are very challenging cases that existing prior methods can hardly handle.

%% file: iclr2026_conference.bib
@inproceedings{ge2016robust,
  title={Robust 3d hand pose estimation in single depth images: from single-view cnn to multi-view cnns},
  author={Ge, Liuhao and Liang, Hui and Yuan, Junsong and Thalmann, Daniel},
  booktitle={Proceedings of the IEEE conference on computer vision and pattern recognition},
  pages={3593--3601},
  year={2016}
}

@inproceedings{oikonomidis2011efficient,
  title={Efficient model-based 3d tracking of hand articulations using kinect},
  author={Oikonomidis, Iason and Kyriazis, Nikolaos and Argyros, Antonis},
  booktitle={Procedings of the British Machine Vision Conference 2011},
  year={2011}
}

@inproceedings{tagliasacchi2015robust,
  title={Robust articulated-icp for real-time hand tracking},
  author={Tagliasacchi, Andrea and Schr{\"o}der, Matthias and Tkach, Anastasia and Bouaziz, Sofien and Botsch, Mario and Pauly, Mark},
  booktitle={Computer graphics forum},
  volume={34-5},
  pages={101--114},
  year={2015},
  organization={Wiley Online Library}
}

@inproceedings{rogez20143d,
  title={3D hand pose detection in egocentric RGB-D images},
  author={Rogez, Gr{\'e}gory and Khademi, Maryam and Supan{\v{c}}i{\v{c}} III, JS and Montiel, Jose Maria Martinez and Ramanan, Deva},
  booktitle={European Conference on Computer Vision},
  pages={356--371},
  year={2014},
  organization={Springer}
}

@inproceedings{simon2017hand,
  title={Hand keypoint detection in single images using multiview bootstrapping},
  author={Simon, Tomas and Joo, Hanbyul and Matthews, Iain and Sheikh, Yaser},
  booktitle={Proceedings of the IEEE conference on Computer Vision and Pattern Recognition},
  pages={1145--1153},
  year={2017}
}

@inproceedings{sridhar2016real,
  title={Real-time joint tracking of a hand manipulating an object from rgb-d input},
  author={Sridhar, Srinath and Mueller, Franziska and Zollh{\"o}fer, Michael and Casas, Dan and Oulasvirta, Antti and Theobalt, Christian},
  booktitle={European conference on computer vision},
  pages={294--310},
  year={2016},
  organization={Springer}
}

@inproceedings{sun2015cascaded,
  title={Cascaded hand pose regression},
  author={Sun, Xiao and Wei, Yichen and Liang, Shuang and Tang, Xiaoou and Sun, Jian},
  booktitle={Proceedings of the IEEE conference on computer vision and pattern recognition},
  pages={824--832},
  year={2015}
}

@article{tompson2014real,
  title={Real-time continuous pose recovery of human hands using convolutional networks},
  author={Tompson, Jonathan and Stein, Murphy and Lecun, Yann and Perlin, Ken},
  journal={ACM Transactions on Graphics (ToG)},
  volume={33},
  number={5},
  pages={1--10},
  year={2014},
  publisher={ACM New York, NY, USA}
}

@article{romero2017embodied,
  title={Embodied hands: modeling and capturing hands and bodies together},
  author={Romero, Javier and Tzionas, Dimitrios and Black, Michael J},
  journal={ACM Transactions on Graphics (TOG)},
  volume={36},
  number={6},
  pages={1--17},
  year={2017},
  publisher={ACM New York, NY, USA}
}

@inproceedings{boukhayma20193d,
  title={3d hand shape and pose from images in the wild},
  author={Boukhayma, Adnane and Bem, Rodrigo de and Torr, Philip HS},
  booktitle={Proceedings of the IEEE/CVF conference on computer vision and pattern recognition},
  pages={10843--10852},
  year={2019}
}

@inproceedings{baek2019pushing,
  title={Pushing the envelope for rgb-based dense 3d hand pose estimation via neural rendering},
  author={Baek, Seungryul and Kim, Kwang In and Kim, Tae-Kyun},
  booktitle={Proceedings of the IEEE/CVF conference on computer vision and pattern recognition},
  pages={1067--1076},
  year={2019}
}

@inproceedings{potamias2023handy,
  title={Handy: Towards a high fidelity 3d hand shape and appearance model},
  author={Potamias, Rolandos Alexandros and Ploumpis, Stylianos and Moschoglou, Stylianos and Triantafyllou, Vasileios and Zafeiriou, Stefanos},
  booktitle={Proceedings of the IEEE/CVF Conference on Computer Vision and Pattern Recognition},
  pages={4670--4680},
  year={2023}
}

@inproceedings{baek2020weakly,
  title={Weakly-supervised domain adaptation via gan and mesh model for estimating 3d hand poses interacting objects},
  author={Baek, Seungryul and Kim, Kwang In and Kim, Tae-Kyun},
  booktitle={Proceedings of the IEEE/CVF conference on computer vision and pattern recognition},
  pages={6121--6131},
  year={2020}
}

@article{kulon2019single,
  title={Single image 3D hand reconstruction with mesh convolutions},
  author={Kulon, Dominik and Wang, Haoyang and G{\"u}ler, Riza Alp and Bronstein, Michael and Zafeiriou, Stefanos},
  journal={arXiv preprint arXiv:1905.01326},
  year={2019}
}

@inproceedings{choi2020pose2mesh,
  title={Pose2mesh: Graph convolutional network for 3d human pose and mesh recovery from a 2d human pose},
  author={Choi, Hongsuk and Moon, Gyeongsik and Lee, Kyoung Mu},
  booktitle={European Conference on Computer Vision},
  pages={769--787},
  year={2020},
  organization={Springer}
}

@inproceedings{chen2021camera,
  title={Camera-space hand mesh recovery via semantic aggregation and adaptive 2d-1d registration},
  author={Chen, Xingyu and Liu, Yufeng and Ma, Chongyang and Chang, Jianlong and Wang, Huayan and Chen, Tian and Guo, Xiaoyan and Wan, Pengfei and Zheng, Wen},
  booktitle={Proceedings of the IEEE/CVF conference on computer vision and pattern recognition},
  pages={13274--13283},
  year={2021}
}

@inproceedings{chen2022mobrecon,
  title={Mobrecon: Mobile-friendly hand mesh reconstruction from monocular image},
  author={Chen, Xingyu and Liu, Yufeng and Dong, Yajiao and Zhang, Xiong and Ma, Chongyang and Xiong, Yanmin and Zhang, Yuan and Guo, Xiaoyan},
  booktitle={Proceedings of the IEEE/CVF conference on computer vision and pattern recognition},
  pages={20544--20554},
  year={2022}
}

@inproceedings{park2022handoccnet,
  title={Handoccnet: Occlusion-robust 3d hand mesh estimation network},
  author={Park, JoonKyu and Oh, Yeonguk and Moon, Gyeongsik and Choi, Hongsuk and Lee, Kyoung Mu},
  booktitle={Proceedings of the IEEE/CVF conference on computer vision and pattern recognition},
  pages={1496--1505},
  year={2022}
}

@inproceedings{ge20193d,
  title={3d hand shape and pose estimation from a single rgb image},
  author={Ge, Liuhao and Ren, Zhou and Li, Yuncheng and Xue, Zehao and Wang, Yingying and Cai, Jianfei and Yuan, Junsong},
  booktitle={Proceedings of the IEEE/CVF conference on computer vision and pattern recognition},
  pages={10833--10842},
  year={2019}
}

@inproceedings{kulon2020weakly,
  title={Weakly-supervised mesh-convolutional hand reconstruction in the wild},
  author={Kulon, Dominik and Guler, Riza Alp and Kokkinos, Iasonas and Bronstein, Michael M and Zafeiriou, Stefanos},
  booktitle={Proceedings of the IEEE/CVF conference on computer vision and pattern recognition},
  pages={4990--5000},
  year={2020}
}

@inproceedings{moon2020i2l,
  title={I2l-meshnet: Image-to-lixel prediction network for accurate 3d human pose and mesh estimation from a single rgb image},
  author={Moon, Gyeongsik and Lee, Kyoung Mu},
  booktitle={European Conference on Computer Vision},
  pages={752--768},
  year={2020},
  organization={Springer}
}

@inproceedings{iqbal2018hand,
  title={Hand pose estimation via latent 2.5 d heatmap regression},
  author={Iqbal, Umar and Molchanov, Pavlo and Gall, Thomas Breuel Juergen and Kautz, Jan},
  booktitle={Proceedings of the European conference on computer vision (ECCV)},
  pages={118--134},
  year={2018}
}

@inproceedings{oh2023recovering,
  title={Recovering 3D hand mesh sequence from a single blurry image: A new dataset and temporal unfolding},
  author={Oh, Yeonguk and Park, JoonKyu and Kim, Jaeha and Moon, Gyeongsik and Lee, Kyoung Mu},
  booktitle={Proceedings of the IEEE/CVF Conference on Computer Vision and Pattern Recognition},
  pages={554--563},
  year={2023}
}

@inproceedings{jiang2023probabilistic,
  title={A probabilistic attention model with occlusion-aware texture regression for 3d hand reconstruction from a single rgb image},
  author={Jiang, Zheheng and Rahmani, Hossein and Black, Sue and Williams, Bryan M},
  booktitle={Proceedings of the IEEE/CVF conference on computer vision and pattern recognition},
  pages={758--767},
  year={2023}
}

@inproceedings{pavlakos2024reconstructing,
  title={Reconstructing hands in 3d with transformers},
  author={Pavlakos, Georgios and Shan, Dandan and Radosavovic, Ilija and Kanazawa, Angjoo and Fouhey, David and Malik, Jitendra},
  booktitle={Proceedings of the IEEE/CVF Conference on Computer Vision and Pattern Recognition},
  pages={9826--9836},
  year={2024}
}

@article{dong2024hamba,
  title={Hamba: Single-view 3d hand reconstruction with graph-guided bi-scanning mamba},
  author={Dong, Haoye and Chharia, Aviral and Gou, Wenbo and Vicente Carrasco, Francisco and De la Torre, Fernando D},
  journal={Advances in Neural Information Processing Systems},
  volume={37},
  pages={2127--2160},
  year={2024}
}

@inproceedings{kim2023sampling,
  title={Sampling is matter: Point-guided 3d human mesh reconstruction},
  author={Kim, Jeonghwan and Gwon, Mi-Gyeong and Park, Hyunwoo and Kwon, Hyukmin and Um, Gi-Mun and Kim, Wonjun},
  booktitle={Proceedings of the IEEE/CVF Conference on computer vision and pattern recognition},
  pages={12880--12889},
  year={2023}
}

@inproceedings{lin2021end,
  title={End-to-end human pose and mesh reconstruction with transformers},
  author={Lin, Kevin and Wang, Lijuan and Liu, Zicheng},
  booktitle={Proceedings of the IEEE/CVF conference on computer vision and pattern recognition},
  pages={1954--1963},
  year={2021}
}

@inproceedings{fan2023arctic,
  title={ARCTIC: A dataset for dexterous bimanual hand-object manipulation},
  author={Fan, Zicong and Taheri, Omid and Tzionas, Dimitrios and Kocabas, Muhammed and Kaufmann, Manuel and Black, Michael J and Hilliges, Otmar},
  booktitle={Proceedings of the IEEE/CVF conference on computer vision and pattern recognition},
  pages={12943--12954},
  year={2023}
}

@inproceedings{zhang2025hawor,
  title={HaWoR: World-space hand motion reconstruction from egocentric videos},
  author={Zhang, Jinglei and Deng, Jiankang and Ma, Chao and Potamias, Rolandos Alexandros},
  booktitle={Proceedings of the Computer Vision and Pattern Recognition Conference},
  pages={1805--1815},
  year={2025}
}

@inproceedings{prakash20243d,
  title={3d hand pose estimation in everyday egocentric images},
  author={Prakash, Aditya and Tu, Ruisen and Chang, Matthew and Gupta, Saurabh},
  booktitle={European Conference on Computer Vision},
  pages={183--202},
  year={2024},
  organization={Springer}
}

@inproceedings{NEURIPS2020_1457c0d6,
 author = {Brown, Tom and Mann, Benjamin and Ryder, Nick and Subbiah, Melanie and Kaplan, Jared D and Dhariwal, Prafulla and Neelakantan, Arvind and Shyam, Pranav and Sastry, Girish and Askell, Amanda and Agarwal, Sandhini and Herbert-Voss, Ariel and Krueger, Gretchen and Henighan, Tom and Child, Rewon and Ramesh, Aditya and Ziegler, Daniel and Wu, Jeffrey and Winter, Clemens and Hesse, Chris and Chen, Mark and Sigler, Eric and Litwin, Mateusz and Gray, Scott and Chess, Benjamin and Clark, Jack and Berner, Christopher and McCandlish, Sam and Radford, Alec and Sutskever, Ilya and Amodei, Dario},
 booktitle = {Advances in Neural Information Processing Systems},
 editor = {H. Larochelle and M. Ranzato and R. Hadsell and M.F. Balcan and H. Lin},
 pages = {1877--1901},
 publisher = {Curran Associates, Inc.},
 title = {Language Models are Few-Shot Learners},
 volume = {33},
 year = {2020}
}

@InProceedings{pmlr-v139-radford21a,
  title = 	 {Learning Transferable Visual Models From Natural Language Supervision},
  author =       {Radford, Alec and Kim, Jong Wook and Hallacy, Chris and Ramesh, Aditya and Goh, Gabriel and Agarwal, Sandhini and Sastry, Girish and Askell, Amanda and Mishkin, Pamela and Clark, Jack and Krueger, Gretchen and Sutskever, Ilya},
  booktitle = 	 {Proceedings of the 38th International Conference on Machine Learning},
  pages = 	 {8748--8763},
  year = 	 {2021},
  editor = 	 {Meila, Marina and Zhang, Tong},
  volume = 	 {139},
  series = 	 {Proceedings of Machine Learning Research},
  month = 	 {18--24 Jul},
  publisher =    {PMLR},
  url = 	 {https://proceedings.mlr.press/v139/radford21a.html}
}

@inproceedings{rubin-etal-2022-learning,
    title = "Learning To Retrieve Prompts for In-Context Learning",
    author = "Rubin, Ohad  and
      Herzig, Jonathan  and
      Berant, Jonathan",
    editor = "Carpuat, Marine  and
      de Marneffe, Marie-Catherine  and
      Meza Ruiz, Ivan Vladimir",
    booktitle = "Proceedings of the 2022 Conference of the North American Chapter of the Association for Computational Linguistics: Human Language Technologies",
    month = jul,
    year = "2022",
    address = "Seattle, United States",
    publisher = "Association for Computational Linguistics",
    url = "https://aclanthology.org/2022.naacl-main.191/",
    doi = "10.18653/v1/2022.naacl-main.191",
    pages = "2655--2671"
}

@article{xie2021explanation,
  title={An explanation of in-context learning as implicit bayesian inference},
  author={Xie, Sang Michael and Raghunathan, Aditi and Liang, Percy and Ma, Tengyu},
  journal={arXiv preprint arXiv:2111.02080},
  year={2021}
}

@article{fang2023explore,
  title={Explore in-context learning for 3d point cloud understanding},
  author={Fang, Zhongbin and Li, Xiangtai and Li, Xia and Buhmann, Joachim M and Loy, Chen Change and Liu, Mengyuan},
  journal={Advances in Neural Information Processing Systems},
  volume={36},
  pages={42382--42395},
  year={2023}
}

@article{liu2025human,
  title={Human-in-Context: Unified Cross-Domain 3D Human Motion Modeling via In-Context Learning},
  author={Liu, Mengyuan and Wang, Xinshun and Fang, Zhongbin and Ye, Deheng and Li, Xia and Tang, Tao and Wu, Songtao and Li, Xiangtai and Yang, Ming-Hsuan},
  journal={arXiv preprint arXiv:2508.10897},
  year={2025}
}

@article{fujii2025towards,
  title={Towards Predicting Any Human Trajectory In Context},
  author={Fujii, Ryo and Saito, Hideo and Hachiuma, Ryo},
  journal={arXiv preprint arXiv:2506.00871},
  year={2025}
}

@inproceedings{johnson2016perceptual,
  title={Perceptual losses for real-time style transfer and super-resolution},
  author={Johnson, Justin and Alahi, Alexandre and Fei-Fei, Li},
  booktitle={European conference on computer vision},
  pages={694--711},
  year={2016},
  organization={Springer}
}

@article{zong2024vlicl,
  title={VL-ICL Bench: The Devil in the Details of Multimodal In-Context Learning},
  author={Zong, Yongshuo and Bohdal, Ondrej and Hospedales, Timothy},
  journal={arXiv preprint arXiv:2403.13164},
  year={2024}
}

@article{zhang2025contextdrivingincontextlearning,
      title={ConText: Driving In-context Learning for Text Removal and Segmentation}, 
      author={Fei Zhang and Pei Zhang and Baosong Yang and Fei Huang and Yanfeng Wang and Ya Zhang},
      year={2025},
      eprint={2506.03799},
      archivePrefix={arXiv},
      primaryClass={cs.CV},
      journal={https://arxiv.org/abs/2506.03799}, 
}

@inproceedings{zhou2023uni3d,
  title={Uni3d: Exploring unified 3d representation at scale},
  author={Zhou, Junsheng and Wang, Jinsheng and Ma, Baorui and Liu, Yu-Shen and Huang, Tiejun and Wang, Xinlong},
  booktitle={International Conference on Learning Representations (ICLR)},
  year={2024}
}

@inproceedings{li2024visual,
  title={Visual in-context prompting},
  author={Li, Feng and Jiang, Qing and Zhang, Hao and Ren, Tianhe and Liu, Shilong and Zou, Xueyan and Xu, Huaizhe and Li, Hongyang and Yang, Jianwei and Li, Chunyuan and others},
  booktitle={Proceedings of the IEEE/CVF Conference on Computer Vision and Pattern Recognition},
  pages={12861--12871},
  year={2024}
}

@inproceedings{Hara2025EventEgoHands,
  author={Hara, Ryosei and Ikeda, Wataru and Hatano, Masashi and Isogawa, Mariko},
  title={EventEgoHands: Event-based Egocentric 3D Hand Mesh Reconstruction},
  booktitle={IEEE International Conference on Image Processing (ICIP)},
  year={2025},
  pages={1199-1204},
  doi={10.1109/ICIP55913.2025.11084751},
}

@inproceedings{he2022masked,
  title={Masked autoencoders are scalable vision learners},
  author={He, Kaiming and Chen, Xinlei and Xie, Saining and Li, Yanghao and Doll{\'a}r, Piotr and Girshick, Ross},
  booktitle={Proceedings of the IEEE/CVF conference on computer vision and pattern recognition},
  pages={16000--16009},
  year={2022}
}

@inproceedings{potamias2025wilor,
  title={Wilor: End-to-end 3d hand localization and reconstruction in-the-wild},
  author={Potamias, Rolandos Alexandros and Zhang, Jinglei and Deng, Jiankang and Zafeiriou, Stefanos},
  booktitle={Proceedings of the Computer Vision and Pattern Recognition Conference},
  pages={12242--12254},
  year={2025}
}

@inproceedings{liu-etal-2022-makes,
    title = "What Makes Good In-Context Examples for {GPT}-3?",
    author = "Liu, Jiachang  and
      Shen, Dinghan  and
      Zhang, Yizhe  and
      Dolan, Bill  and
      Carin, Lawrence  and
      Chen, Weizhu",
    editor = "Agirre, Eneko  and
      Apidianaki, Marianna  and
      Vuli{\'c}, Ivan",
    booktitle = "Proceedings of Deep Learning Inside Out (DeeLIO 2022): The 3rd Workshop on Knowledge Extraction and Integration for Deep Learning Architectures",
    month = may,
    year = "2022",
    address = "Dublin, Ireland and Online",
    publisher = "Association for Computational Linguistics",
    url = "https://aclanthology.org/2022.deelio-1.10/",
    doi = "10.18653/v1/2022.deelio-1.10",
    pages = "100--114"
}

@inproceedings{NEURIPS2023_398ae57e,
 author = {Zhang, Yuanhan and Zhou, Kaiyang and Liu, Ziwei},
 booktitle = {Advances in Neural Information Processing Systems},
 editor = {A. Oh and T. Naumann and A. Globerson and K. Saenko and M. Hardt and S. Levine},
 pages = {17773--17794},
 publisher = {Curran Associates, Inc.},
 title = {What Makes Good Examples for Visual In-Context Learning?},
 volume = {36},
 year = {2023}
}

@article{bar2022visual,
  title={Visual prompting via image inpainting},
  author={Bar, Amir and Gandelsman, Yossi and Darrell, Trevor and Globerson, Amir and Efros, Alexei},
  journal={Advances in Neural Information Processing Systems},
  volume={35},
  pages={25005--25017},
  year={2022}
}

@inproceedings{wang2023images,
  title={Images speak in images: A generalist painter for in-context visual learning},
  author={Wang, Xinlong and Wang, Wen and Cao, Yue and Shen, Chunhua and Huang, Tiejun},
  booktitle={Proceedings of the IEEE/CVF Conference on Computer Vision and Pattern Recognition},
  pages={6830--6839},
  year={2023}
}

@misc{bai2023qwentechnicalreport,
      title={Qwen Technical Report}, 
      author={Jinze Bai and Shuai Bai and Yunfei Chu and Zeyu Cui and Kai Dang and Xiaodong Deng and Yang Fan and Wenbin Ge and Yu Han and Fei Huang and Binyuan Hui and Luo Ji and Mei Li and Junyang Lin and Runji Lin and Dayiheng Liu and Gao Liu and Chengqiang Lu and Keming Lu and Jianxin Ma and Rui Men and Xingzhang Ren and Xuancheng Ren and Chuanqi Tan and Sinan Tan and Jianhong Tu and Peng Wang and Shijie Wang and Wei Wang and Shengguang Wu and Benfeng Xu and Jin Xu and An Yang and Hao Yang and Jian Yang and Shusheng Yang and Yang Yao and Bowen Yu and Hongyi Yuan and Zheng Yuan and Jianwei Zhang and Xingxuan Zhang and Yichang Zhang and Zhenru Zhang and Chang Zhou and Jingren Zhou and Xiaohuan Zhou and Tianhang Zhu},
      year={2023},
      eprint={2309.16609},
      archivePrefix={arXiv},
      primaryClass={cs.CL},
      url={https://arxiv.org/abs/2309.16609}, 
}

@article{qwen2.5,
    title   = {Qwen2.5 Technical Report}, 
    author  = {An Yang and Baosong Yang and Beichen Zhang and Binyuan Hui and Bo Zheng and Bowen Yu and Chengyuan Li and Dayiheng Liu and Fei Huang and Haoran Wei and Huan Lin and Jian Yang and Jianhong Tu and Jianwei Zhang and Jianxin Yang and Jiaxi Yang and Jingren Zhou and Junyang Lin and Kai Dang and Keming Lu and Keqin Bao and Kexin Yang and Le Yu and Mei Li and Mingfeng Xue and Pei Zhang and Qin Zhu and Rui Men and Runji Lin and Tianhao Li and Tingyu Xia and Xingzhang Ren and Xuancheng Ren and Yang Fan and Yang Su and Yichang Zhang and Yu Wan and Yuqiong Liu and Zeyu Cui and Zhenru Zhang and Zihan Qiu},
    journal = {arXiv preprint arXiv:2412.15115},
    year    = {2024}
}

@misc{
loshchilov2018fixing,
title={Fixing Weight Decay Regularization in Adam},
author={Ilya, Frank Hutter},
year={2018},
url={https://openreview.net/forum?id=rk6qdGgCZ},
}

@inproceedings{grauman2024ego,
  title={Ego-exo4d: Understanding skilled human activity from first-and third-person perspectives},
  author={Grauman, Kristen and Westbury, Andrew and Torresani, Lorenzo and Kitani, Kris and Malik, Jitendra and Afouras, Triantafyllos and Ashutosh, Kumar and Baiyya, Vijay and Bansal, Siddhant and Boote, Bikram and others},
  booktitle={Proceedings of the IEEE/CVF Conference on Computer Vision and Pattern Recognition},
  pages={19383--19400},
  year={2024}
}

@inproceedings{fan2021learning,
  title={Learning to disambiguate strongly interacting hands via probabilistic per-pixel part segmentation},
  author={Fan, Zicong and Spurr, Adrian and Kocabas, Muhammed and Tang, Siyu and Black, Michael J and Hilliges, Otmar},
  booktitle={2021 International Conference on 3D Vision (3DV)},
  pages={1--10},
  year={2021},
  organization={IEEE}
}

@inproceedings{moon2020interhand2,
  title={Interhand2. 6m: A dataset and baseline for 3d interacting hand pose estimation from a single rgb image},
  author={Moon, Gyeongsik and Yu, Shoou-I and Wen, He and Shiratori, Takaaki and Lee, Kyoung Mu},
  booktitle={European Conference on Computer Vision},
  pages={548--564},
  year={2020},
  organization={Springer}
}

@inproceedings{zheng2023potter,
    title={POTTER: Pooling Attention Transformer for Efficient Human Mesh Recovery},
    author={Zheng, Ce and Liu, Xianpeng and Qi, Guo-Jun and Chen, Chen},
    booktitle={Proceedings of the IEEE/CVF Conference on Computer Vision and Pattern Recognition},
    year={2023}
}

@misc{chen2024pcieegohandposesolutionegoexo4dhand,
      title={PCIE EgoHandPose Solution for EgoExo4D Hand Pose Challenge}, 
      author={Feng Chen and Ling Ding and Kanokphan Lertniphonphan and Jian Li and Kaer Huang and Zhepeng Wang},
      year={2024},
      eprint={2406.12219},
      archivePrefix={arXiv},
      primaryClass={cs.CV},
      url={https://arxiv.org/abs/2406.12219}, 
}

@inproceedings{
xu2025do,
title={Do Egocentric Video-Language Models Truly Understand Hand-Object Interactions?},
author={Boshen Xu and Ziheng Wang and Yang Du and Zhinan Song and Sipeng Zheng and Qin Jin},
booktitle={The Thirteenth International Conference on Learning Representations},
year={2025},
url={https://openreview.net/forum?id=M8gXSFGkn2}
}

@InProceedings{Yang_2025_CVPR,
    author    = {Yang, Jingkang and Liu, Shuai and Guo, Hongming and Dong, Yuhao and Zhang, Xiamengwei and Zhang, Sicheng and Wang, Pengyun and Zhou, Zitang and Xie, Binzhu and Wang, Ziyue and Ouyang, Bei and Lin, Zhengyu and Cominelli, Marco and Cai, Zhongang and Li, Bo and Zhang, Yuanhan and Zhang, Peiyuan and Hong, Fangzhou and Widmer, Joerg and Gringoli, Francesco and Yang, Lei and Liu, Ziwei},
    title     = {EgoLife: Towards Egocentric Life Assistant},
    booktitle = {Proceedings of the IEEE/CVF Conference on Computer Vision and Pattern Recognition (CVPR)},
    month     = {June},
    year      = {2025},
    pages     = {28885-28900}
}

@article{dosovitskiy2020image,
  title={An image is worth 16x16 words: Transformers for image recognition at scale},
  author={Dosovitskiy, Alexey and Beyer, Lucas and Kolesnikov, Alexander and Weissenborn, Dirk and Zhai, Xiaohua and Unterthiner, Thomas and Dehghani, Mostafa and Minderer, Matthias and Heigold, Georg and Gelly, Sylvain and others},
  journal={arXiv preprint arXiv:2010.11929},
  year={2020}
}

@article{vaswani2017attention,
  title={Attention is all you need},
  author={Vaswani, Ashish and Shazeer, Noam and Parmar, Niki and Uszkoreit, Jakob and Jones, Llion and Gomez, Aidan N and Kaiser, {\L}ukasz and Polosukhin, Illia},
  journal={Advances in neural information processing systems},
  volume={30},
  year={2017}
}

@article{li2024llava,
  	title={LLaVA-OneVision: Easy Visual Task Transfer},
  	author={Li, Bo and Zhang, Yuanhan and Guo, Dong and Zhang, Renrui and Li, Feng and Zhang, Hao and Zhang, Kaichen and Li, Yanwei and Liu, Ziwei and Li, Chunyuan},
  	journal={arXiv preprint arXiv:2408.03326},
  	year={2024}
}

@article{dong2022survey,
  title={A survey on in-context learning},
  author={Dong, Qingxiu and Li, Lei and Dai, Damai and Zheng, Ce and Ma, Jingyuan and Li, Rui and Xia, Heming and Xu, Jingjing and Wu, Zhiyong and Liu, Tianyu and others},
  journal={arXiv preprint arXiv:2301.00234},
  year={2022}
}

@article{ferber2024context,
  title={In-context learning enables multimodal large language models to classify cancer pathology images},
  author={Ferber, Dyke and W{\"o}lflein, Georg and Wiest, Isabella C and Ligero, Marta and Sainath, Srividhya and Ghaffari Laleh, Narmin and El Nahhas, Omar SM and M{\"u}ller-Franzes, Gustav and J{\"a}ger, Dirk and Truhn, Daniel and others},
  journal={Nature Communications},
  volume={15},
  number={1},
  pages={10104},
  year={2024},
  publisher={Nature Publishing Group UK London}
}

@InProceedings{Kwon_2021_ICCV,
    author    = {Kwon, Taein and Tekin, Bugra and St\"uhmer, Jan and Bogo, Federica and Pollefeys, Marc},
    title     = {H2O: Two Hands Manipulating Objects for First Person Interaction Recognition},
    booktitle = {Proceedings of the IEEE/CVF International Conference on Computer Vision (ICCV)},
    month     = {October},
    year      = {2021},
    pages     = {10138-10148}
}

@article{hong2022pointcam,
  title={Pointcam: Cut-and-mix for open-set point cloud learning},
  author={Hong, Jie and Qiu, Shi and Li, Weihao and Anwar, Saeed and Harandi, Mehrtash and Barnes, Nick and Petersson, Lars},
  journal={arXiv preprint arXiv:2212.02011},
  year={2022}
}

@inproceedings{zhu2024ssp,
  title={Ssp: Semi-signed prioritized neural fitting for surface reconstruction from unoriented point clouds},
  author={Zhu, Runsong and Kang, Di and Hui, Ka-Hei and Qian, Yue and Qiu, Shi and Dong, Zhen and Bao, Linchao and Heng, Pheng-Ann and Fu, Chi-Wing},
  booktitle={Proceedings of the IEEE/CVF Winter Conference on Applications of Computer Vision},
  pages={3769--3778},
  year={2024}
}

@article{cui2023adaptive,
  title={Adaptive low rank adaptation of segment anything to salient object detection},
  author={Cui, Ruikai and He, Siyuan and Qiu, Shi},
  journal={arXiv preprint arXiv:2308.05426},
  year={2023}
}

@article{cui2024numgrad,
  title={Numgrad-pull: Numerical gradient guided tri-plane representation for surface reconstruction from point clouds},
  author={Cui, Ruikai and Xie, Binzhu and Qiu, Shi and Liu, Jiawei and Anwar, Saeed and Barnes, Nick},
  journal={arXiv preprint arXiv:2411.17392},
  year={2024}
}
